\documentclass[10pt,twocolumn,twoside]{IEEEtran}

\usepackage{CJKutf8}
\usepackage[normalem]{ulem}
\usepackage{cite}
\usepackage{url}
\usepackage{graphicx}
\usepackage{booktabs}
\usepackage{tabularx}
\usepackage{multirow}
\usepackage{subfigure}
\usepackage{amsmath,dsfont}
\usepackage{array}
\usepackage{subfigure}
\usepackage{mathtools}

\usepackage{algorithm,algorithmic}

\usepackage{verbatim}
\usepackage{bbm}

\usepackage{colortbl}
\definecolor{mygray}{gray}{.75}

\usepackage{scalerel,stackengine}

\usepackage{framed}
\usepackage{color}
\usepackage{xcolor}
\usepackage{amssymb}

\newcommand{\red}[1]{\textcolor{black}{#1}}
\newcommand{\blue}[1]{\textcolor{black}{#1}}
\usepackage[normalem]{ulem}

\newcommand{\eg}{\emph{e.g.,}~}
\newcommand{\etal}{\emph{et al.}~}
\newcommand{\ie}{\emph{i.e.,}~}

\newcommand{\xr}[1]{\textcolor{purple}{[\textbf{XR}: #1]}}
\newcommand{\resnext}{ResNeXt-101}

\newcommand{\specialcell}[2][l]{%
  \begin{tabular}[#1]{@{}l@{}}#2\end{tabular}}

{\begin{snugshade}\begin{quote}}
{\hfill\end{quote}\end{snugshade}}

\definecolor{shadecolor}{rgb}{0.9,0.9,0.9}

\begin{document}
\begin{CJK*}{UTF8}{gbsn}

\title{COCO-CN for Cross-Lingual Image Tagging, Captioning and Retrieval}

\author{Xirong~Li,~Chaoxi Xu,~Xiaoxu Wang,~Weiyu Lan,~Zhengxiong Jia,~Gang Yang,~Jieping Xu
\thanks{Manuscript received July 01, 2018; revised October 27, 2018 and December 25, 2018; accepted January 13, 2019. 

This work was supported by NSFC (No. 61672523, No. 61771468), the Fundamental Research Funds for the Central Universities and the Research Funds of Renmin University of China (No. 18XNLG19).
The associate editor coordinating the review of this manuscript and approving it for publication was Dr Wolfgang H\"urst.
(\textit{Corresponding author: Jieping Xu}).}

\thanks{The authors are with the Key Lab of Data Engineering and Knowledge Engineering, Renmin University of China, and the AI \& Media Computing Lab, School of Information, Renmin University of China, Beijing 100872, China (e-mail: \{xirong,xjieping\}@ruc.edu.cn) }
}

\markboth{IEEE Transactions on Multimedia ,~Vol.~?, No.~?, ?~2019}%
{Li \MakeLowercase{\textit{et al.}}:COCO-CN for Cross-Lingual Image Tagging, Captioning and Retrieval}

\maketitle

\begin{abstract}
This paper contributes to cross-lingual image annotation and retrieval in terms of data and baseline methods. We propose \emph{COCO-CN}, a novel dataset enriching MS-COCO with manually written Chinese sentences and tags. For more effective annotation acquisition, we develop a recommendation-assisted collective annotation system, automatically providing an annotator with several tags and sentences deemed to be relevant with respect to the pictorial content. Having 20,342 images annotated with 27,218 Chinese sentences and 70,993 tags, COCO-CN is currently the largest Chinese-English dataset \blue{that provides a unified and challenging platform} for cross-lingual image tagging, captioning and retrieval. We develop conceptually simple yet effective methods per task for learning from cross-lingual resources.  Extensive experiments on the \red{three} tasks justify the viability of the proposed dataset and methods. \blue{Data and code are publicly available at \textcolor{blue}{\url{https://github.com/li-xirong/coco-cn}}}.
\end{abstract}


\IEEEpeerreviewmaketitle

\section{Introduction} \label{sect:intro}

\IEEEPARstart{A}{utomated} description of multimedia content, let it be image  or video, is among the core themes for multimedia analysis and retrieval. For a long time, efforts on describing multimedia have limited the form of description to be individual words \cite{tmm18-cui}, \eg \emph{car}, \emph{road}, and \emph{people}, due to the semantic gap between low-level features and high-level semantics. Only recently we have witnessed noticeable advances in producing natural language-like descriptions, \eg \emph{cars racing on a road surrounded by lots of people}, that indicate not only the presence of specific concepts but also their interactions \cite{tmm17-gao}. While these advances can certainly be attributed to technical breakthroughs, mostly deep learning, the importance of well-labeled datasets shall not be underestimated. Exemplars are ImageNet \cite{deng2009imagenet} for visual class recognition, NUS-WIDE \cite{chua2009nus} for image tagging, MS-COCO \cite{chen2015microsoft} for image captioning, and more recently Twitter100k~\cite{tmm18-twitter100k} for cross-media retrieval, to name a few. Notice that annotations of these datasets are all written in the English language. To facilitate research that goes beyond the monolingual setting, this paper presents a new dataset called \emph{COCO-CN}. It extends MS-COCO with manually written Chinese sentences and tags, see Table \ref{tab:coco-cn}. 

\begin{table}[tb!]
\renewcommand{\arraystretch}{1.1}
\centering
\caption{\textbf{Examples from the proposed COCO-CN dataset}. The new dataset extends MS-COCO~\cite{chen2015microsoft} with manually written Chinese sentences and tags. Text in parentheses is English translation, provided for non-Chinese readers.}
\label{tab:coco-cn} 
\scalebox{0.8}{
\begin{tabular}{@{}c lll@{}}
\toprule
               & & \multicolumn{2}{c}{\textbf{COCO-CN}} \\
               \cmidrule{3-4} 
\textbf{Image} & \textbf{MS-COCO text} &  \emph{Sentences} & \emph{Tags} \\
\midrule
 \multirow{6}{*}{\includegraphics[width=2.05cm,height=2.05cm]{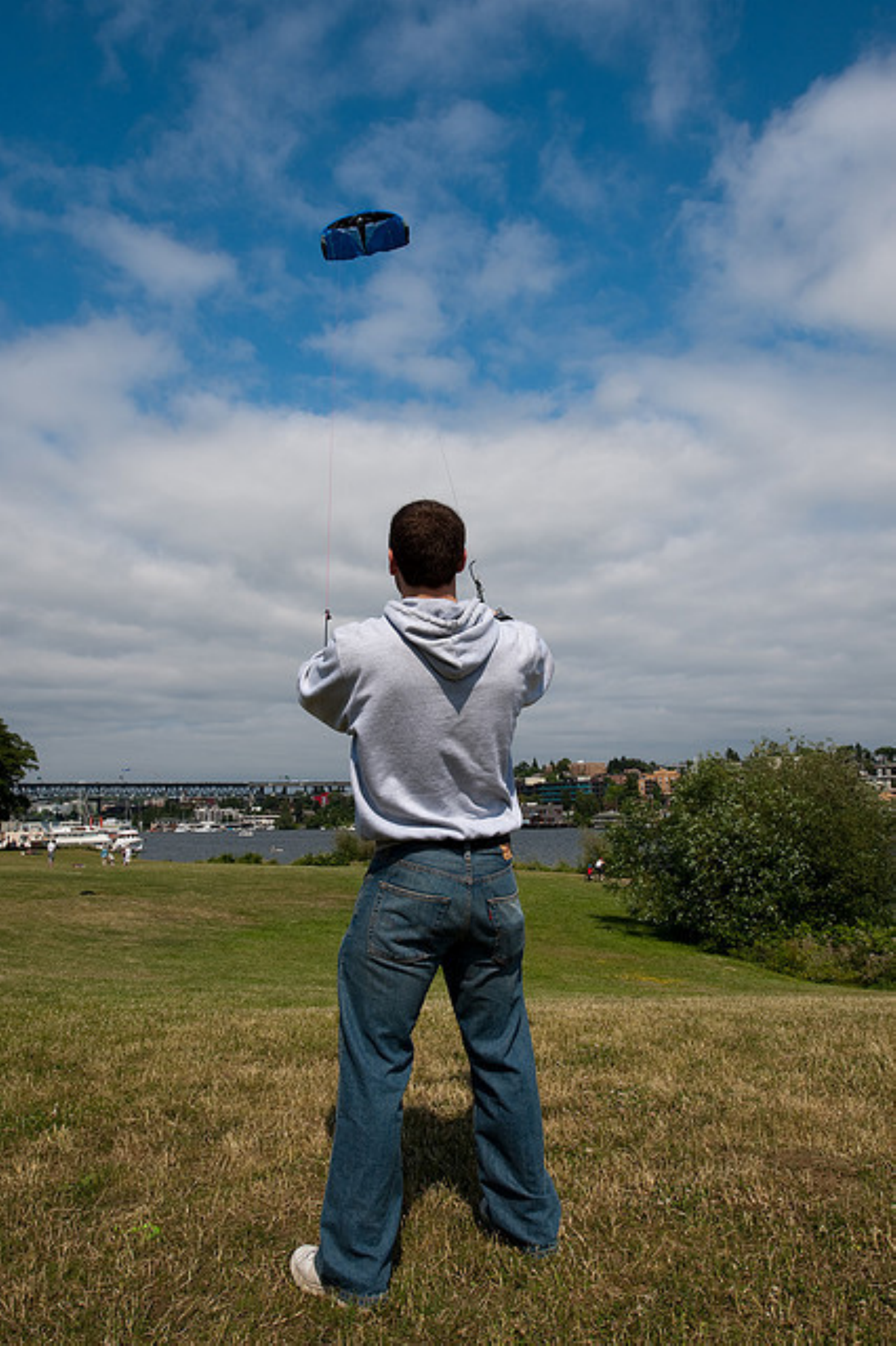}} & \multirow{6}{0.25\columnwidth}{\specialcell{man is flying a \\ kite in a field}} & \multirow{6}{0.4\columnwidth}{\specialcell{一个男人在河边的 \\ 草地上放风筝\\(A man flying a kite \\ on the riverside grass)}} & \multirow{6}{0.15\columnwidth}{\specialcell{蓝天~(bule sky) \\ 天空~(sky) \\ 多云~(cloudy) \\ 年轻人~(youth) \\ 青草~(grass) \\ 草地~(lawn)}} \\
 & \\
 & \\
 & \\
 & \\
 & \\
 \cmidrule{2-4}
 \multirow{8}{*}{\includegraphics[width=2.05cm,height=2.05cm]{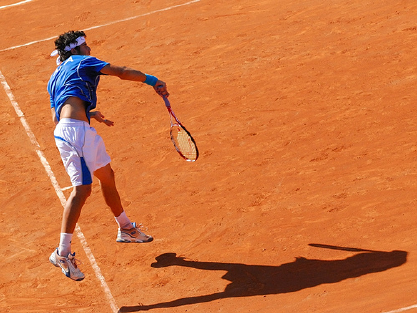}} & \multirow{7}{0.25\columnwidth}{\specialcell{young man \\ serving a tennis \\ ball to  his \\ opponent}} & \multirow{7}{0.4\columnwidth}{\specialcell{在红土场上，一个\\身穿蓝色运动服的\\男人正跳起发球\\(On a clay court, \\a man wearing a \\ blue sportswear is \\ jumping to serve)}} & \multirow{6}{0.15\columnwidth}{\specialcell{红土球场~(clay\\court) \\ 网球~(tennis) \\ 男人~(man) \\ 球拍~(racket) \\ 网球运动员~(tennis\\player) \\ 红土~(red clay)}} \\ 
 & \\
 & \\
 & \\
 & \\
 & \\ 
 & \\ 
 & \\
 \cmidrule{2-4}
 \multirow{5}{*}{\includegraphics[width=2.05cm,height=2.05cm]{COCO_train2014_000000571964}} & \multirow{5}{0.25\columnwidth}{\specialcell{zebras run on \\ the grass near \\the trees}} & \multirow{5}{0.4\columnwidth}{\specialcell{一群野生动物在\\草地上奔跑~(A herd \\ of wild animals are \\ running  on the grass)}} & \multirow{5}{0.15\columnwidth}{\specialcell{灌木丛~(bush) \\ 斑马~(zebra) \\ 草地~(grass) \\大自然~(nature)}} \\
 & \\
 & \\
 & \\
 & \\
 & \\

\bottomrule
\end{tabular}
}
\end{table}

Although the importance of multimedia annotation and retrieval in a \emph{cross-lingual} setting has been recognized early~\cite{iapr-tc12}, only recently has the topic gained increasing attention \cite{multilingual-caption,elliott2016multi30k,miyazaki2016cross,hitschler2016multimodal, rajendran2015bridge,cbmi17qijie,weiyu2017fluency,ijcnlp17Elliott}. \blue{Cross-lingual image tagging is important as it studies how to exploit data labeled in a source language to obtain image tagging models for a target language. Consequently, common users are provided with the possibility to have their images automatically tagged in a native language even when training examples in that language are in short supply}. In \cite{miyazaki2016cross}, for instance, Miyazaki and Shimizu exploit a transfer learning strategy, initializing the visual embedding matrix of a Japanese captioning model using its counterpart from a trained English captioning model. Tsutsui and Crandall propose to generate multilingual sentences in a single network, using artificial tokens to control the language \cite{multilingual-caption}. Lan \etal demonstrate the possibility of training an image captioning model by exploiting English resources to generate fluent Chinese sentences \cite{weiyu2017fluency}.  Elliott \etal show that using images as a side information improves multi-modal machine translation \cite{ijcnlp17Elliott}. Cross-lingual applications are crucial for the majority of the world's population who do not speak English. 

To support this line of research, several image datasets with non-English annotations have been developed. Depending on their applications, the target languages of these datasets differ, including German for multi-modal machine translation \cite{hitschler2016multimodal,elliott2016multi30k}, German and French for image retrieval \cite{rajendran2015bridge}, Japanese for cross-lingual document retrieval \cite{funaki2015image} and image captioning \cite{miyazaki2016cross,yoshikawa2017stair}, and Chinese for image captioning \cite{flickr8kcn,weiyu2017fluency,wu2017ai}. Except for IAPR TC-12 \cite{iapr-tc12} and AIC-ICC \cite{wu2017ai} which consist of images collected from the Internet, each of these datasets is built \red{on top} of an existing English dataset, with MS-COCO as the most popular choice. 
 
Despite the encouraging progress, we see two deficiencies in the current research in terms of datasets and methods. While Chinese is the most spoken language in the world, existing datasets for this language are either small in scale (Flickr8k-CN \cite{flickr8kcn}) or strongly biased for describing human activities and monolingual (AIC-ICC \cite{wu2017ai}). As for methods, there is a paucity of literature on cross-lingual image tagging. Initially set up for cross-lingual purposes, the ImageCLEF series have factually become monolingual \cite{imageclef2013}. The only work published in the last five years\footnote{Paper \cite{cbmi17qijie} was the only hit when querying Google Scholar with ``cross-lingual image annotation'' or ``cross-lingual image tagging''.} is by Wei \etal \cite{cbmi17qijie}, where the authors assume zero-availability of training images in their target language. Clarifai \cite{clarifai}, an online image tagging service, is capable of predicting tags in multiple languages. But the non-English tags appear to be simply machine translation of the predicted English tags. \emph{How to effectively perform cross-lingual image tagging is unclear}.

Works on cross-lingual image captioning consistently report that for training an image captioning model for a target language, machine-translated sentences are less effective than manually written sentences ~\cite{miyazaki2016cross,yoshikawa2017stair,weiyu2017fluency,flickr8kcn}. No attempt is made to answer \emph{whether and in what ways machine-translated data can be effectively exploited together with manually-written sentences in the target language} for better performance. For matching images and sentences in two different languages, Gella \etal propose a neural model to learn multilingual multi-modal representations with image as pivot \cite{emnlp17gella}. A shortcoming of that model is it requires many triplets for training, where each image is associated with bilingual descriptions. Cross-lingual image retrieval requires not only cross-modal matching between a query and an image, but also cross-lingual matching between the same query and the description associated with the image, but
in another language.

\begin{figure}[tb!]
\centering
\includegraphics[width=\columnwidth]{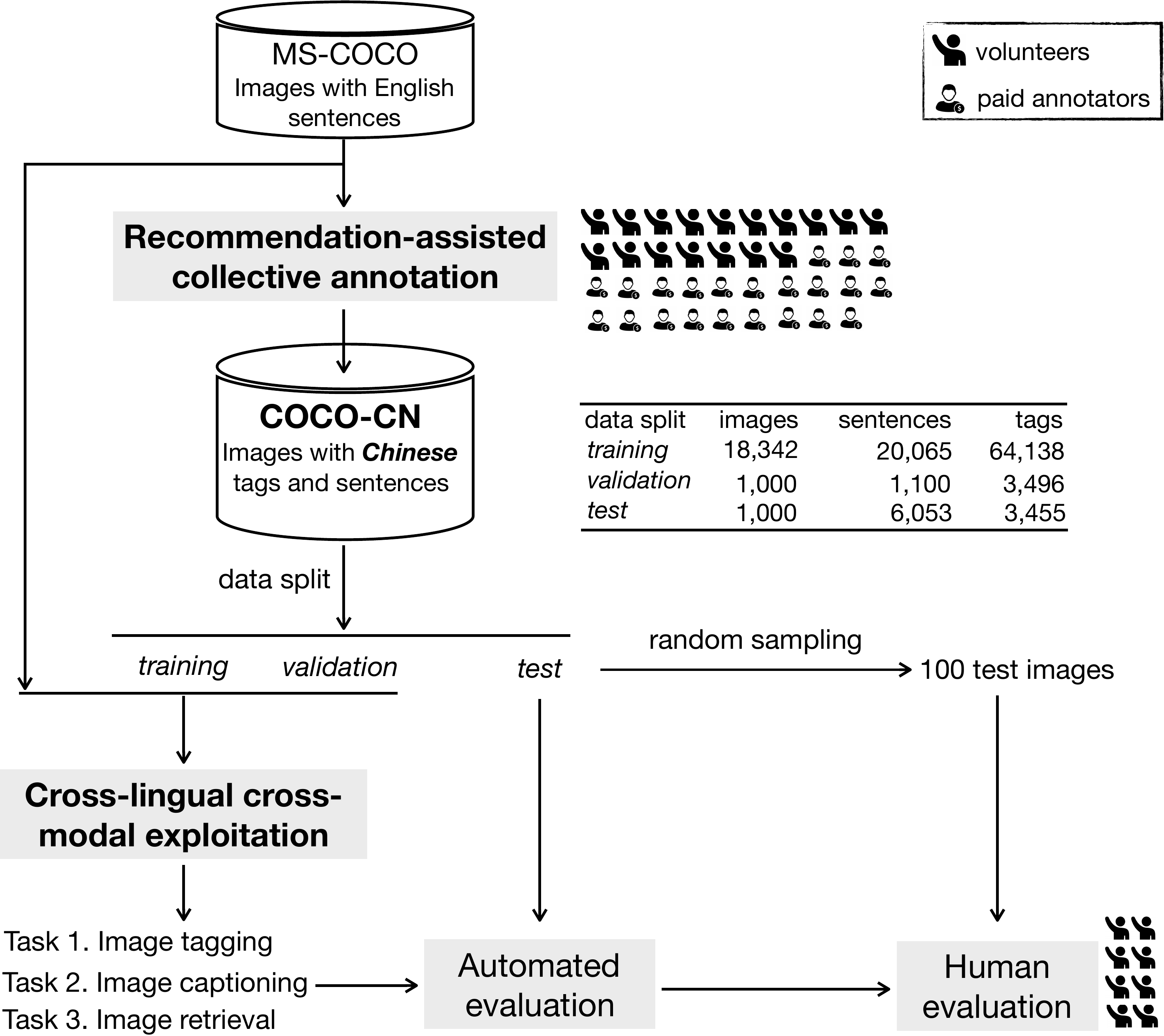}
\caption{\textbf{The development of COCO-CN and its usage in three tasks}.}
\label{fig:framework}
\end{figure}

To attack the above two deficiencies, this paper makes three contributions as follows: \\ [2pt]
$\bullet$ We present \emph{COCO-CN}, a new dataset extending MS-COCO with manually written Chinese sentences and tags. It thus differs from previous cross-lingual efforts on MS-COCO that target Japanese \cite{miyazaki2016cross,yoshikawa2017stair}, German \cite{hitschler2016multimodal} or French \cite{rajendran2015bridge}. With over 20k images described and tagged, COCO-CN is the largest Chinese-English dataset for cross-lingual image annotation and retrieval. The development of COCO-CN and its usage in the varied tasks are illustrated in Fig. \ref{fig:framework}. Data \blue{and code are} available at \url{https://github.com/li-xirong/coco-cn}. \\ [2pt]
$\bullet$ For COCO-CN construction a novel recommendation-assisted annotation system is developed. While much progress has been made for image auto-tagging \cite{li16-tagsurvey,fu_rui_2017,tmm18-lixue} and caption retrieval \cite{tcsvt2017-cross-media-reivew,tmm18-lzhang,w2vv}, whether these techniques can be used to assist dataset construction is unexplored. The new system allows us to assess the possibility of exploring the state-of-the-art for reducing annotation workload. \blue{Moreover, with its source code released, the system provides a good starting point for peers who are interested in building similar datasets}. \\ [2pt]
$\bullet$ We show the applicability of COCO-CN for three tasks, \ie cross-lingual image tagging, captioning and retrieval, and its superiority over alternate datasets. Moreover, we develop \blue{models specifically for the cross-lingual setting}, \ie \emph{Cascading MLP} for tagging, \emph{Sequential Learning} for captioning and \emph{Enhanced Word2VisualVec} for retrieval. While conceptually simple, these models effectively learn from \blue{bilingual} resources and consequently advance the state-of-the-art in the cross-lingual \blue{context}.

\begin{table*}[tb!]
\renewcommand{\arraystretch}{1.1}
\centering
\caption{\textbf{Public image datasets with manually annotated, non-English descriptions}. With over 20k images described and tagged, COCO-CN is currently the largest Chinese-English dataset applicable for cross-lingual image tagging, captioning and retrieval.}
\label{tab:datasets}
\scalebox{0.9}{
\begin{tabular}{@{}lrllrrll@{}}
\toprule
\textbf{Dataset} & \textbf{Release} & \textbf{Data source} & \textbf{Languages} & \textbf{Images} & \textbf{Sentences} & \textbf{Tags} & \textbf{Applications}\\
\midrule
IAPR TC-12 \cite{iapr-tc12} & 2006 & Internet & English / German & 20,000 & 70,460 & no & image retrieval \\
Japanese Pascal sentences \cite{funaki2015image} & 2015 & Pascal sentences \cite{rashtchian2010collecting} & Japanese / English  & 1,000 & 5,000 & no & cross-lingual document retrieval \\

YJ Captions \cite{miyazaki2016cross} & 2016 & MS-COCO \cite{chen2015microsoft} & Japanese / English & 26,500 & 131,740 & no & image captioning \\

STAIR Captions \cite{yoshikawa2017stair} & 2017 & MS-COCO & Japanese / English & 164,062 & 820,310 & no & image captioning \\

MIC test data \cite{rajendran2015bridge} & 2016 & MS-COCO & French / German / English & 1,000 & 5,000 & no & image retrieval \\

Bilingual caption \cite{hitschler2016multimodal} & 2016 & MS-COCO & German / English & 1,000 & 1,000 & no & machine translation / image captioning \\

Multi30k \cite{elliott2016multi30k} & 2016 & Flickr30k \cite{flickr30k} & German / English & 31,014 & 186,084 & no & machine translation / image captioning \\

Flickr8k-CN \cite{flickr8kcn} & 2016 & Flickr8k~\cite{flickr8k} & Chinese / English & 8,000 & 45,000 & no & image captioning \\

AIC-ICC \cite{wu2017ai} &  2017 & Internet & Chinese & 240,000 & 1,200,000 & no & image captioning \\

Flickr30k-CN \cite{weiyu2017fluency} & 2017 & Flickr30k & Chinese / English & 1,000 & 5,000 & no & image captioning \\ 

\emph{COCO-CN (this paper)}  & 2018 & MS-COCO & Chinese / English & 20,342 & 27,218 & yes & image tagging / captioning / retrieval \\ 
\bottomrule
\end{tabular}
}
\end{table*}

\section{Related Work} \label{sect:related}

We summarize in Table \ref{tab:datasets} an (incomplete) list of publicly available image datasets that are associated with manually annotated, non-English descriptions. The datasets closely related to ours in terms of the target language are Flickr8k-CN \cite{flickr8kcn}, Flickr30k-CN \cite{weiyu2017fluency} and AIC-ICC \cite{wu2017ai}, all focusing on Chinese.

To the best of our knowledge, Flickr8k-CN is the first dataset for image captioning in Chinese \cite{flickr8kcn}, adopting Flickr8k \cite{flickr8k} as its data source. Each image \red{has been} re-annotated with five Chinese sentences written by native speakers via a local crowd sourcing service. Probably because no annotation guideline or quality control \red{was} provided, the Flickr8k-CN descriptions tend to be short, containing on average 8 Chinese characters per sentence. As for Flickr30k-CN \cite{weiyu2017fluency}, its manual annotation covers only the test set of Flickr30k~\cite{flickr30k}, by collectively translating 5,000 test sentences from English to Chinese. AIC-ICC is a big step forward, consisting of 240k images collected from Internet search engines and 1.2 million crowd-sourced sentences \cite{wu2017ai}. With an average number of 21.6 Chinese characters per sentence, the texts appear to be more descriptive than their Flickr8k-CN counterpart. Nevertheless, AIC-ICC has two downsides. First, it is a monolingual dataset, inapplicable in a cross-lingual setting. Second, all its images are about human activities, requiring human figures to be clearly visible. Thereby the AIC-ICC sentences are tailored to human activity description. As a consequence, an image captioning model trained on AIC-ICC \blue{always} generates sentences with a pattern of someone doing something, even if no person is present in test images.

COCO-CN naturally inherits the visual diversity of MS-COCO. Its cross-lingual applicability is granted by jointly exploiting the new Chinese and the existing English annotations. Moreover, we collect tags to further enrich the annotations. \blue{The tags can be used for multi-label image classification. They can also be applied to multi-modal tasks, \eg image captioning with tags.} All this makes COCO-CN a versatile dataset for cross-lingual image tagging, captioning and retrieval.

\section{COCO-CN Construction} \label{sect:dataset}

The complexity of pictorial content makes it non-trivial to write a fluent sentence to fully describe an image. Consider the image in the first row of Table \ref{tab:coco-cn} for instance. The gist of the scene is a man flying a kite outdoor. If one tries to cover the entire context, \eg sky, cloud, river, tree, and grass field, it will be a lengthy sentence that does not read smoothly. 
So we collect tags to complement natural-language descriptions.

\subsection{Recommendation-assisted Collective Annotation}

\subsubsection{User Interface}

\begin{figure*}[tb!]
\centering
\includegraphics[width=1.9\columnwidth]{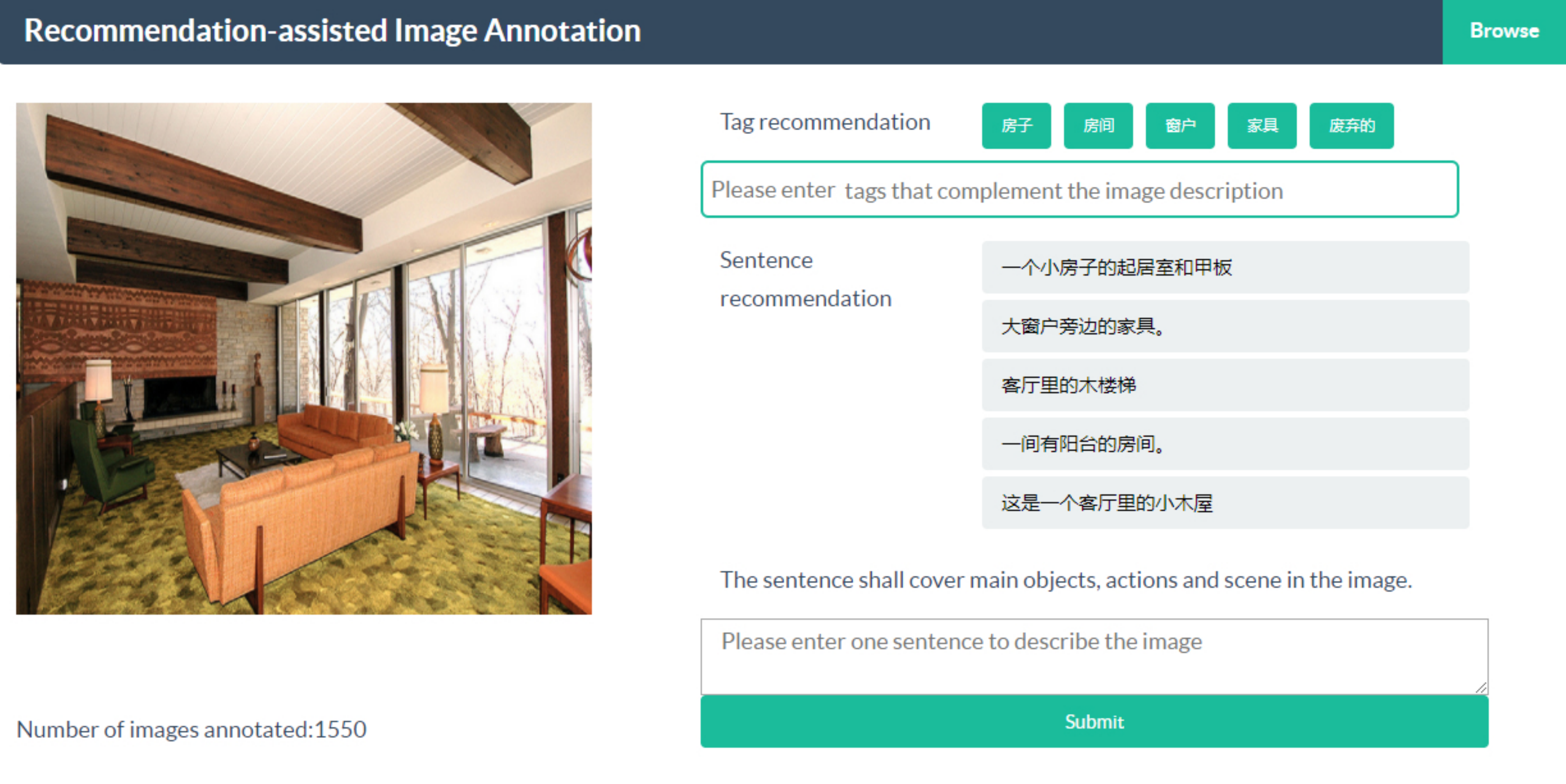}
\caption{\textbf{A snapshot of our web based image annotation system}. Rather than starting from scratch, an annotator is provided with five tags and five sentences automatically recommended by the system based on the pictorial content. When clicked, the tags and sentences will instantly appear in the editable text forms to assist manual annotation. \blue{Source code of the annotation system is publicly available at https://tinyurl.com/cococn-system}.}
\label{fig:system}
\end{figure*}

For the ease of participation, we develop a web based image annotation system that allows a user to annotate images remotely and independently. Fig. \ref{fig:system} provides a snapshot of the system. Our novel design is the deployment of two content-based recommendation modules, one for sentences and the other for tags, that assist manual annotation on the fly. When clicked, a sentence or tag from the recommendation board will instantly appear in the corresponding editable text form. User operations including clicking and editing are logged for data analytics.

Since there is no off-the-shelf model available for recommending Chinese sentences or tags based on the pictorial content, we build our own models, as described in Section \ref{sssec:sent-recom} and \ref{sssec:tag-recom}. Notice that we started the COCO-CN project in 2016 with the different modules developed in different periods, so the choice of image features was subject to the availability of the corresponding CNN models.

\subsubsection{Chinese Sentence Recommendation} \label{sssec:sent-recom}

Given an image to be annotated, our sentence recommendation module retrieves the sentence most likely to be relevant from a large Chinese sentence pool. The pool, consisting of 120k sentences, is obtained by performing machine translation on the MS-COCO English sentences. Li \etal make an empirical comparison between two public machine translation services, \ie Google and Baidu, reporting the better performance of Baidu for translating image descriptions from English to Chinese~\cite{flickr8kcn}. So we use Baidu translation.

Given that each image is associated with five machine-translated sentences, a straightforward strategy would be to simply allow the annotators to select the most appropriate one and edit it as they see fit. We do not choose this strategy for the following considerations. First, we encourage the annotators, while being assisted, to describe the images from their (personal) viewpoints, rather than just make a good translation. Learning from images with unaligned bilingual descriptions is more challenging yet more practical.  Second, we want to keep the system flexible to handle novel images without any descriptions. Last but not least, to justify the usability of sentence retrieval for manual annotation.

For sentence retrieval we adopt Word2VisualVec (W2VV) \cite{w2vv}, which demonstrates state-of-the-art performance on multiple datasets. W2VV is designed to predict deep visual features from just an (English) textual input. In particular, the mapping from a given text to a visual feature vector is achieved by first encoding the text into a fixed-length vector. The vector then goes through a Multilayer Perceptron (MLP) to produce the visual feature vector. Consequently, the cross-modal relevance between a given image and a given sentence is computed as cosine similarity in the visual feature space. While W2VV is originally developed for retrieving English captions, it can be easily extended to the Chinese scenario by substituting its English based encoding layer for a Chinese counterpart. Paired images and Chinese sentences are required to train the Chinese version of W2VV. To that end, we split at random the pool of 120k sentences into two disjoint subsets, with 90\% for training and 10\% for validation. We train a W2VV model that uses the bag-of-words based encoding and predicts a 2,048-dim ResNet-152 feature. As shown in Fig. \ref{fig:system}, for a given image we compute its relevance to each sentence in the pool on the fly, and recommend the top-5 retrieved sentences to an annotator.

\subsubsection{Chinese Tag Recommendation} \label{sssec:tag-recom}

Our tag recommendation module is to predict multiple Chinese tags based on the visual content. Since the MS-COCO images are quite diverse, a large vocabulary is required. We need a large-scale training set wherein each image is labeled with multiple Chinese tags. Since a dataset as such is not publicly available, we consider Flickr images associated with user tags in different languages. However, we observe that the \red{number} of Flickr images labeled with Chinese tags is much less than its English counterpart. This is in line with \cite{koochali2016flickrlang}, where Koochali \etal report that the majority of the Flickr tags are in English. Also we note that the quality of Chinese tags is much lower. Therefore, we use Flickr images but manually translate the tag vocabulary from English to Chinese. In particular, we adopt one million Flickr images provided by \cite{li16-tagsurvey}. To construct a Chinese tag vocabulary, we first sort all the English tags in the 1M set in descending order by the \red{number} of distinct users who have used them for image tagging. We manually go through the ranked tag list, performing one-by-one manual translation. Tags lacking  correspondence to specific objects, scenes or events are excluded. Note that synonyms are grouped. For instance, `car', `auto' and `automobile' are all translated into the Chinese word `汽车'. Consequently, images labeled with `car', `auto' or `automobile' are considered as positive training examples of `汽车'. A word with multiple meanings is translated to a list of Chinese words. For instance, `glass' is mapped to \{`玻璃', `玻璃杯', `眼镜'\}. We let the annotators decide which translation is the most appropriate. Finally we obtain a Chinese vocabulary of 1,951 classes, covering 2,085 English and 2,367 Chinese words. 

As Flickr data is known to be noisy~\cite{li16-tagsurvey,tmm18-cui,tmm18-jlda}, we de-noise with the following rule of thumb. We remove overly-tagged images (associated with more than 20 tags) and images having less than two tags from the constructed vocabulary. This results in a training set of 800k images. 
For image representation we employ a pre-trained GoogLeNet model \cite{mettes2016imagenet}, taking the output of its pool5 layer as a 1,024-dim visual feature. An MLP with a network structure of 1024-1024-2048-1951 is trained on the 800k images. 
The top-5 Chinese tags predicted by the MLP model \red{are} presented to each annotator.


\subsection{COCO-CN Annotation Process} \label{ssec:anno-process}

Sentences of existing datasets such as MS-COCO, STAIR captions, Flickr8k-CN and AIC-ICC were gathered by distributing the annotation task on a specific crowd-sourcing platform. However, how to measure and thus control the quality of crowd-sourced sentences remains unresolved. Consider MS-COCO for instance. During manual translation we found a number of typos and misspellings in its sentences, \eg ``ready to hi~[hit] a ball'', ``smile son [on] their faces'', ``airplane with a striped tale [tail]'', and ``a giraffe standing by a palm team [tree]''. In order to gather high-quality annotations, we did not fully rely on crowd sourcing. Our annotation team was comprised of 17 volunteers (staff and graduate students in our lab) and 22 paid undergraduate students from our department. 

Before they started, the team were instructed with the following guidelines. A sentence shall cover \red{the} main objects, actions and scene in a given image.  The annotators were asked to provide at least one tag that complements the image description. For a better understanding of the guidelines, we provided a number of well / badly labeled examples. For quality control, an inspection committee comprised of paper authors regularly sampled a small proportion of the annotation results by distinct annotators for \red{a} manual check. Annotators responsible for unqualified results were put into a watch list and informed to review and re-edit their annotations. An annotator entering the watch list three times would be excluded from the team, with his/her annotations removed as well. 

For effective acquisition of labeled data, our goal is to have more images annotated than to collect more sentences per image. This strategy is inspired by ~\cite{w2vv} where the authors empirically show that given the same \red{number} of image-sentence pairs for training, having more images results in a better model for cross-modal matching. To that end, we generated for each annotator a unique list of images by randomly shuffling the MS-COCO images. Consequently, 
the chance of two annotators labeling the same image was low.


\subsection{Data Analytics} \label{ssec:data-analysis}

\subsubsection{Progress Overview}

We have 20,342 images annotated with 22,218 sentences and 70,993 tags in total. Notice that we did not select these 20k images in advance. They were obtained as a consequence of manual annotation. Since images were randomly assigned to each annotator, the 20k images in COCO-CN are factually a representative subset of MS-COCO. In addition, we have 5,000 MS-COCO English sentences \emph{manually} translated to Chinese for evaluating image captioning (c.f. Section \ref{ssect:cap}). This adds the \red{number} of manually written Chinese sentences up to 27,218. Different from an English sentence where words are separated by whitespace characters, a Chinese sentence has no marker as word boundaries. It has to be segmented into a sequence of meaningful words before further analysis. We employ the BosonNLP toolkit~\cite{boson} for sentence segmentation and part-of-speech tagging. Basic statistics of the sentences and tags \red{are} listed in Table \ref{tab:Basic_character}. The average number of annotators per image is 1.1, while the average number of images annotated per person is 521.6.

\begin{table}[tb!]
\renewcommand{\arraystretch}{1.2}
\centering
\caption{\textbf{Statistics of the COCO-CN sentences and tags}. The sentences contribute a vocabulary of 7,096 distinct words, where the percentages of nouns, verbs and adjectives are 59.5\%, 22.6\% and 5.1\%. The number of distinct tags is 4,867. The number of tags outside the sentence vocabulary is 1,918.}
\label{tab:Basic_character} 
\scalebox{0.95}{
\begin{tabular}{@{}crrrrrrrrrr@{}}
\toprule
       \multicolumn{3}{c}{\textbf{Chars per sentence}} && \multicolumn{3}{c}{\textbf{Words per sentence}} && \multicolumn{3}{c}{\textbf{Tags per image}}   \\
       \cmidrule{1-3} \cmidrule{5-7} \cmidrule{9-11} 
\textit{min} & \textit{max} & \textit{mean} && \textit{min} & \textit{max} & \textit{mean} && \textit{min} & \textit{max} & \textit{mean}\\
4 & 59 & 16.8 && 2 & 43 & 10.9 && 1 & 15 & 3.5  \\
\bottomrule
\end{tabular}
}
\end{table}

\textbf{Chinese tag vocabulary}. The sentences contribute a vocabulary of 7,096 distinct words, while the manual tags contribute a vocabulary of 4,867 distinct words, among which 1,918 words are outside the sentence vocabulary. At the image level, on average nearly 45\% of the manual tags are not covered by the associated sentences. It is clear that only tags with a reasonable amount of training images can be effectively modeled. So we expand the manual tags with words automatically extracted from Chinese sentences as follows. We consider only nouns, adjectives and verbs. Adjectives and verbs consisting of single Chinese characters, \eg `满', `亮' and `穿', are over generic and thus removed. We further exclude rare tags that \red{occur} less than 20 times in the COCO-CN training set (see Fig. \ref{fig:framework} and Section \ref{ssec:setup} for data partition). We then manually go through the remaining tags, deleting those we consider unsuited for describing the pictorial content. Consequently, we obtain a vocabulary of 655 Chinese tags. The number of positive training images ranges from 20 to 1,876, with an averaged number of 123.

\textbf{COCO-CN tags \textit{versus} MS-COCO object classes}. The images in the MS-COCO dataset are already annotated with 80 object classes such as \emph{person}, \emph{bicycle} and \emph{toothbrush}. After manually translating these classes into Chinese tags, we find that there are 70 tags in common. In other words, we have enriched the MS-COCO annotations with $655-70=585$ novel concepts. 

\textbf{COCO-CN sentences \textit{versus} MS-COCO sentences}. For the ease of cross-lingual comparison, for each image in COCO-CN we employ machine translation to convert its five English captions from MS-COCO to Chinese. Given a COCO-CN sentence we investigate the overlap between its keywords, \ie nouns, verbs and adjectives, and those in the translated sentences. On average a COCO-CN sentence brings in 45.3\% novel keywords, with 39.9\% nouns, 51.9\% verbs, and 69.5\% adjectives. The usage of adjectives is more subjective and personalized. Consider for instance describing a dog as \emph{small} or \emph{big} and judging if a cat is \emph{cute}.


\textbf{Annotation quality}. Concerning the annotation team, no one entered the watch list three times. 
The annotators were asked to perform the annotation task independently, which helped improve the diversity of human annotations. The regular inspection mechanism, with nearly 10\% of the images double-checked, helped maintain the overall annotation quality. Concerning the variety of description for each image and the vocabulary for the description, we evaluate in two aspects as follows. First, for each image we use its five machine translated sentences as references, and compute four metrics, i.e., BLEU-4, METEOR, ROUGE-L and CIDEr, that have been \emph{de facto} for evaluating caption quality. The corresponding scores are 16.1, 27.6, 42.9, and 60.0, respectively, suggesting a substantial divergence between human annotations and machine translations. Second, we compare their vocabulary size. Not surprisingly, the human vocabulary, with \red{a} size of 7,096, is smaller than the machine vocabulary which has 14,219 distinct words. Notice however that the \red{number} of machine-translated sentences is five times as large as the \red{number} of manually written sentences. So the human vocabulary is reasonably diverse. \blue{Moreover, we checked low-level typos in each manually written sentence by matching each of its words with a vocabulary of nearly 350k words from the Jieba Chinese text segmentation tool. For 942 out of the 22,218 sentences, they contain words outside the vocabulary. A manual check shows that only 56 sentences have true typos, \eg incorrectly typing `西兰花'~(broccoli) as `西蓝花' or `牛仔裤'~(jeans) as `牛子裤'. For the other 886 sentences, their mismatched words such as `矮柜' (low cabinet) and `幼象' (baby elephant) are actually valid}. The accuracy of the human annotations will be further verified by image captioning experiments, where COCO-CN is compared against current alternatives including Flickr8k-CN and AIC-ICC as training data.
We also compare statistics of the sentences respectively written by the volunteer group and the paid group, and find no considerable difference.

\subsubsection{The Influence of Recommendation}

We now analyze if the two recommendation modules are helpful during the  annotation process. 

\begin{figure}[tb!]
\centering
\includegraphics[width=0.95\columnwidth]{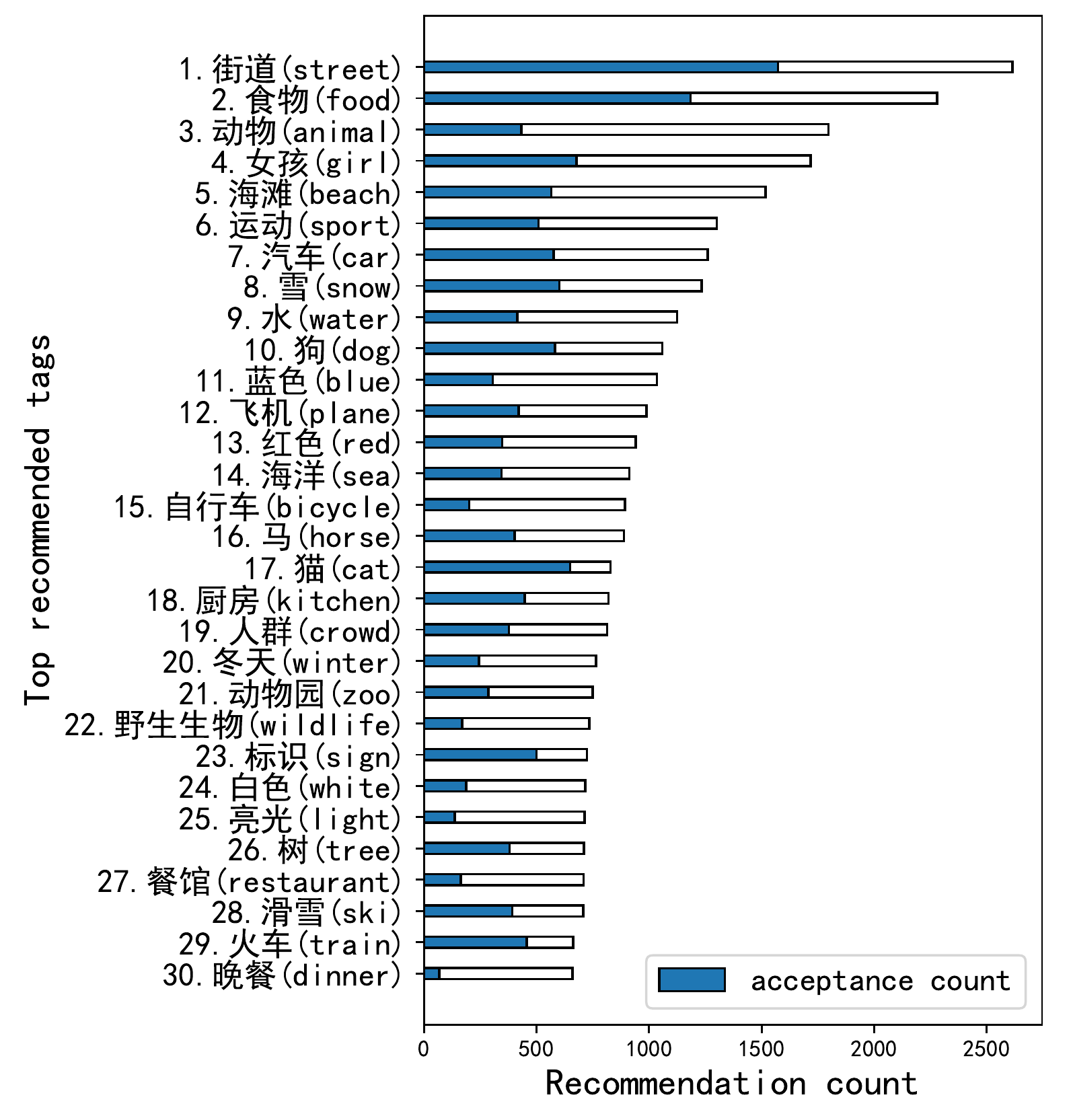}
\caption{\textbf{The top-30 recommended tags}. Their English translations are provided in parentheses for non-Chinese readers.}
\label{fig:toptags}
\end{figure}

\textbf{Tag recommendation}. The annotators accepted approximately 2.7 out of the five suggested tags per image, with an acceptance rate of 54\%. Fig. \ref{fig:toptags} shows the top 30 recommended tags and their acceptance counts. The best\red{-}predicted tag is `猫' (\emph{cat}), got accepted 649 out of 828 times.

\begin{figure}[tb!]
\centering
\includegraphics[width=\columnwidth]{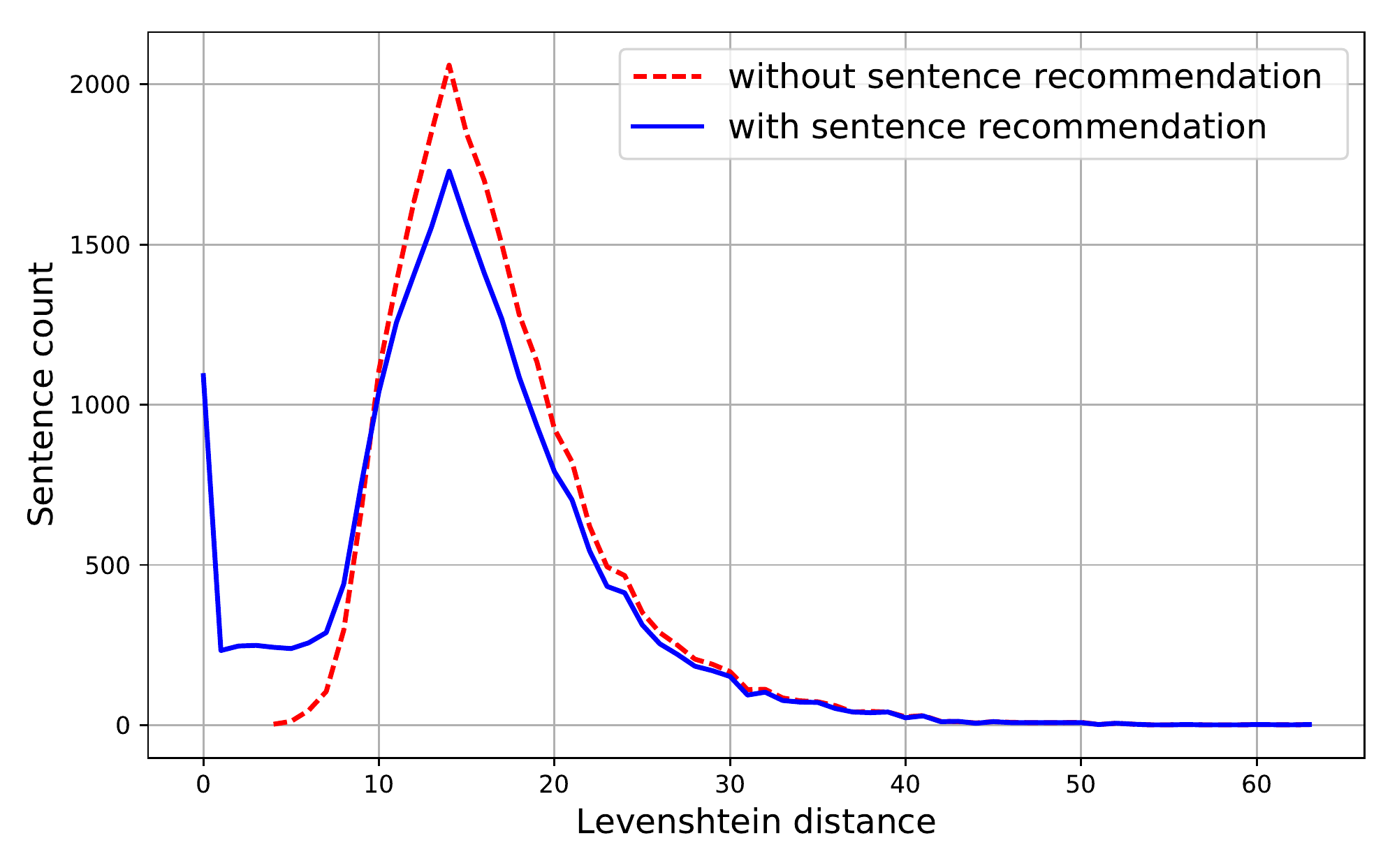}
\caption{\textbf{Distribution of the Levenshtein distance with and without sentence recommendation}. Sentence recommendation decreases the distance, and thus reduces annotation workload.}
\label{fig:LD}
\end{figure}

\textbf{Sentence recommendation}. Among the 22,218 sentences, 31\% of them were typed by the annotators after clicking (and optionally re-editing) the recommendations. The number of edited sentences is 5,796. To measure the manual effort saved per image, we propose to calculate the Levenshtein distance \cite{1966SPhD...10..707L} between the recommended sentence clicked by an annotator and the final submitted sentence. As a string metric, the Levenshtein distance between two Chinese sentences is the minimum number of single Chinese character edits (insertions, deletions or substitutions) required to change one sentence into the other. Our baseline is sentence annotation without recommendation, \blue{where} the distance between the annotation and the recommendation (which is null) is \blue{actually} the length of the annotation. The baseline has an average distance of 16.8, while the recommendation counterpart is 15.0. Fig. \ref{fig:LD} shows the distance distribution curves with and without recommendation. These results suggest that sentence recommendation reduces the annotation workload to some extent. Nevertheless, sentence recommendation is less effective than tag recommendation. 

\section{Applications} \label{sect:app}

\subsection{Common setup} \label{ssec:setup}

Before detailing the individual tasks, we first describe an experimental setup shared by the tasks.

\textbf{Data partition}. We randomly select 1,000 images from COCO-CN as a held-out test set, and another set of 1,000 images as a validation set for hyper parameter optimization. The remaining 18,342 images are used for training. When evaluating image tagging and captioning in an automatic manner, limitations exist due to vocabulary discrepancy and incomplete ground truth. Therefore, we randomly select 100 images from the test set, which we term COCO-CN test100, for human evaluation. In addition, for human evaluation we construct another test set of 100 images, which are randomly sampled from NUS-WIDE other than MS-COCO.

\textbf{Image feature}. For the three tasks, we employ the state-of-the-art ResNeXt-101 model \cite{resnext} with input size of $224\times224$. The model, provided by the MediaMill team at University of Amsterdam, is pre-trained on the full ImageNet dataset with a bottom-up reorganization \cite{mettes2016imagenet} of the 22k ImageNet classes. The reorganization balances the classes by grouping or deleting over-specific and rare classes and down-sampling training examples for over-frequent classes, and consequently reduces the number of classes to 12k. We take the output of the last pooling layer, obtaining a 2,048-dim feature vector. For each image we resize it to $256\times256$, and extract its feature using an oversampling strategy. That is, we extract features from its 10 sub images, obtained by clipping the image and its horizontal flip with a window of $224\times224$ at their center and four corners. The 10 features are averaged as the image feature. Note that the ResNeXt-101 feature is better than the GoogLeNet and ResNet-152 features previously used in our annotation system, see the appendix.

\subsection{Task I: Cross-Lingual Image Tagging} \label{ssect:tagging}
We describe how to exploit the Chinese annotations from COCO-CN and the English annotations from MS-COCO for building an image tagging model that predicts Chinese tags for a given image.


\subsubsection{Setup}

\begin{figure}[tb!]
\centering
\includegraphics[width=\columnwidth]{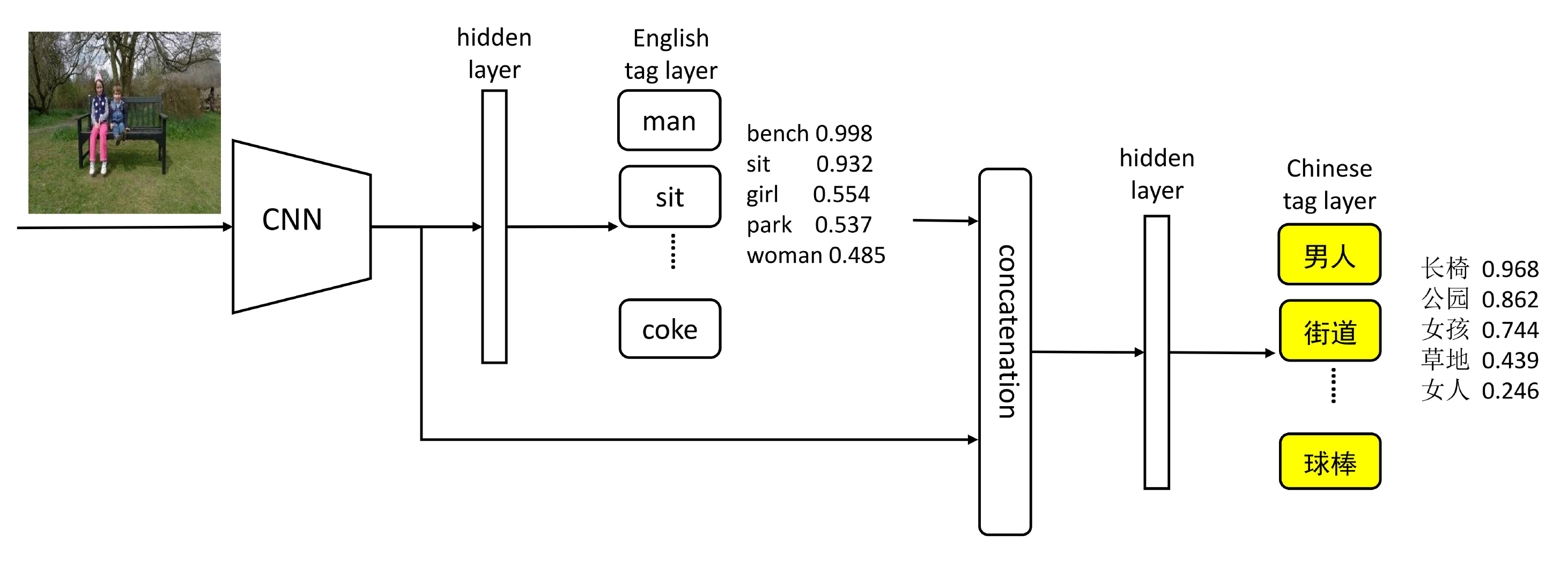}
\caption{\textbf{The proposed Cascading MLP for cross-lingual image tagging}. We train the first MLP that predicts English tags. The output of this MLP is concatenated with the image CNN feature to form the input for the second MLP, which predicts Chinese tags.}
\label{fig:cross-tagging}
\end{figure}

\textbf{Tagging model}. For learning from the cross-lingual resources, we propose Cascading MLP. As illustrated in Fig. \ref{fig:cross-tagging}, the model consists of two MLPs. The first MLP is trained on MS-COCO for English tag prediction. \red{A} vocabulary of 512 English tags is constructed from the original MS-COCO captions by sorting nouns, verbs and adjectives in descending order in terms of their occurrence frequency. As such, this MLP essentially transforms the visual feature of a given image into a 512-dim semantic feature, with each dimension corresponding to a specific English tag. The visual and semantic features are concatenated to form a semantic-enhanced input of the second MLP. This MLP is trained using COCO-CN to predict the probability of each Chinese tag being relevant with respect to a given image. As shown in Fig. \ref{fig:cross-tagging}, each MLP has one hidden layer consisting of 1,024 neurons. A rectified linear unit (ReLU) is followed to increase the nonlinearity of the model. As an image can be described by multiple tags, we use sigmoid (rather than softmax commonly used for visual recognition) as the activation function of the two output layers. \blue{The loss function is binary cross entropy}. During training we apply a common dropout trick (with a rate of 0.5) to reduce the risk of overfitting.

\textbf{Baselines}. In order to justify the effectiveness of the proposed model, we consider the following four alternates. \\
\emph{1) COCO-MT}. An MLP is trained using tags extracted from the machine-translated Chinese sentences of MS-COCO. Leaving out 2k images covered by the COCO-CN validation and test sets, COCO-MT has 121,287 training images. \\
\emph{2) COCO-CN}. An MLP is trained on COCO-CN alone. \\
\emph{3) Clarifai}. An online image tagging service~\cite{clarifai} that predicts for a given image up to 200 tags in multiple languages including Chinese. The service claims to be a market leader since winning the top five places in image classification at the ImageNet 2013 competition, and is accessible via \red{its} API. \\
\emph{4) Multi-task MLP}. A multi-task MLP is trained on COCO-CN with bi-lingual annotations to simultaneously predict English and Chinese tags. The model is a  non­cascaded version of Fig. \ref{fig:cross-tagging}, where the sum of the English and Chinese \blue{binary cross-entropy losses is}  minimized during training.


\textbf{Evaluation criteria}. We report the popular Precision, Recall and F-measure at top 5. Each metric is computed per image and averaged over all test images.

\subsubsection{Results}

\textbf{Automated evaluation}. Table \ref{tab:tagexp} shows automated evaluation results of the \red{five} tagging models. The proposed Cascading MLP tops the performance. Although the number of training images in COCO-MT is 6.6 times as large as COCO-CN, the MLP trained on COCO-CN outperforms its COCO-MT counterpart with a clear margin. This result shows the importance of high-quality annotation for training. Some image tagging results are presented in Table VII. In the first row, the tag \emph{umbrella} is not predicted by MLP trained on COCO-CN, while in the second row, MLP trained on COCO-MT incorrectly predicts the tag \emph{keyboard}. Learning from the two complementary datasets, Cascading MLP makes better prediction in general.

Multi-task MLP is better than MLP trained on COCO-CN, suggesting a positive effect of multi-task regularization. Nonetheless, it is less effective than Cascading MLP. Moreover, in contrast to Multi-task MLP, Cascading MLP, by learning two monolingual MLPs sequentially, does not require aligned bi-lingual annotations per training image, and is thus more flexible to exploit partially annotated yet larger training data. Therefore, the cascading architecture is more suited for learning from \blue{unaligned} cross-lingual resources.

\blue{We also compare Multi-task MLP and Cascading MLP on Flickr8k-CN, where the result shows that the latter is better. For more details we refer to \url{https://tinyurl.com/tag-flickr8kcn}}.

\textbf{Human evaluation}. It is possible that the lower performance of Clarifai is caused by a discrepancy between the Chinese vocabulary of the online service and our ground-truth vocabulary. To resolve this uncertainty, we performed a user study as follows. For each of the 100 pre-specified images, we collected the top 5 predicted tags by each model. Eight subjects participated \red{in} the user study. Each image together with the collected tags \red{was} shown to two subjects, who independently rated each of the tags as \emph{relevant}, \emph{irrelevant} or \emph{unsure}. To avoid bias, the tags were randomly shuffled in advanced. Only tags rated as relevant by both subjects were preserved. Table \ref{tab:tagexp-human} shows the tagging performance given the new ground truth. As the ground truth is more complete, all the scores improve. Cascading MLP again performs the best. Moreover, the qualitative conclusion concerning which model performs better remains the same. 

\begin{table} [tb!]
\renewcommand{\arraystretch}{1.1}
\caption{\textbf{Automated evaluation of different models for image tagging}. Cascading MLP learned from cross-lingual data is the best.}
\label{tab:tagexp} 
\centering
\scalebox{0.9}{
\begin{tabular}{@{}llrrr@{}}
\toprule
\textbf{Model} & \textbf{Precision} & \textbf{Recall}  & \textbf{F-measure}  \\
\midrule
Clarifai~\cite{clarifai}  & 0.217 & 0.261  & 0.228 \\
MLP  trained on COCO-MT & 0.432 & 0.525  & 0.456\\
MLP  trained on COCO-CN & 0.477 & 0.576  & 0.503\\
Multi-task MLP          & 0.482 & 0.583  & 0.508 \\ 
Cascading MLP  & \textbf{0.491} & \textbf{0.594} & \textbf{0.517}\\
\bottomrule
\end{tabular}
} 
\end{table}

\begin{table} [tb!]
\renewcommand{\arraystretch}{1.1}
\caption{\textbf{Human evaluation of different models for image tagging on COCO-CN test100}. Cascading MLP again performs the best.}
\label{tab:tagexp-human} 
\centering
\scalebox{0.9}{
\begin{tabular}{@{}llrrr@{}}
\toprule
\textbf{Model} & \textbf{Precision} & \textbf{Recall}  & \textbf{F-measure}  \\
\midrule
Clarifai  & 0.634 & 0.358  & 0.451 \\
MLP  trained on COCO-MT & 0.778 & 0.453  & 0.563\\
MLP  trained on COCO-CN & 0.836 & 0.488  & 0.607\\
Cascading MLP  & \textbf{0.858} & \textbf{0.501} & \textbf{0.623}\\
\bottomrule
\end{tabular}
} 
\end{table}

To prevent dataset bias, we repeated the human study on another test set of 100 images randomly sampled from NUS-WIDE~\cite{chua2009nus}, which is independent of all the training sets, \ie Flickr8k-CN, AIC-ICC, COCO-MT and COCO-CN, used in this work. We term this additional test set NUS-WIDE100. As  Table \ref{tab:tagexp-human-nuswide100} shows, Cascading MLP is again the best.

\begin{table} [tb!]
\renewcommand{\arraystretch}{1.1}
\caption{\textbf{Human evaluation of image tagging on NUS-WIDE100}, a random subset of 100 images from NUS-WIDE.}
\label{tab:tagexp-human-nuswide100} 
\centering
\scalebox{0.9}{
\begin{tabular}{@{}llrrr@{}}
\toprule
\textbf{Model} & \textbf{Precision} & \textbf{Recall}  & \textbf{F-measure}  \\
\midrule
Clarifai  & 0.636   & 0.448 &   0.515 \\
MLP  trained on COCO-MT & 0.668 & 0.472 & 0.542 \\
MLP  trained on COCO-CN & 0.694 & 0.487 & 0.561 \\
Cascading MLP  &  \textbf{0.726} &   \textbf{0.509} &  \textbf{0.588} \\
\bottomrule
\end{tabular}
} 
\end{table}

\begin{table*}[tb!]
\renewcommand{\arraystretch}{1.2}
\centering
\caption{\textbf{Some tagging and captioning results by different models}. Texts in parentheses are English translations, provided for non-Chinese readers.}
\label{tab:tagging-example} 
\scalebox{0.8}{
\begin{tabular}{@{}c l l@{}}
\toprule
\textbf{Test image}  &  \textbf{Results} \\
\midrule
 \multirow{10}{*}{\includegraphics[width=3cm]{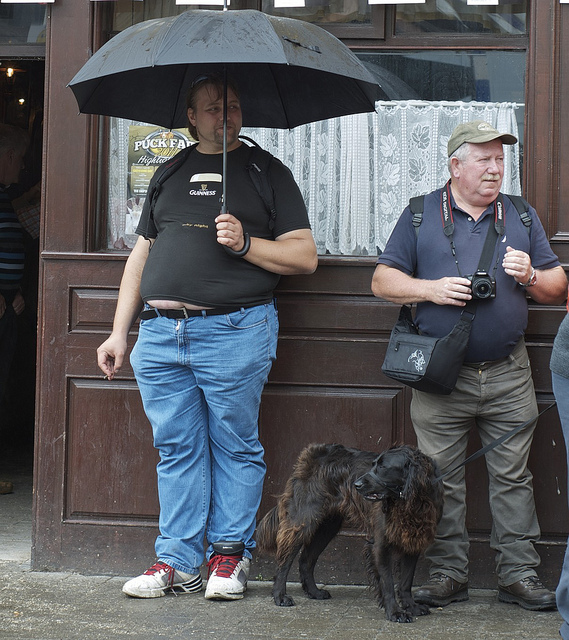}} 
 & \emph{\textbf{Image Tagging}}: \\ 
 & Manual: 雨伞 (umbrella), 帽子 (hat), 男人 (man)  狗 (dog), 相机 (camera), 黑色 (black) \\
 & COCO-MT: 狗 (dog), 男人 (man), 伞 (umbrella), 雨伞 (umbrella), 女人 (woman) \\
 & COCO-CN: 狗 (dog), 男人 (man), 黑色 (black), 女人 (woman), 街道 (street) \\
 & Clarifai: 人 (person), 巡逻 (patrol), 女人 (woman), 男性 (male), 狗 (dog) \\
 & Multi-task MLP: 狗 (dog), 男人 (man), 街道 (street), 女人 (woman), 黑色 (black) \\
 & Cascading MLP: 狗 (dog), 男人 (man), 雨伞 (umbrella), 黑色 (black), 女人 (woman) \\ [1pt]
 & \emph{\textbf{Image Captioning}}: \\
 & Manual: 一个挎着相机的男人旁站着一位撑伞的男子和一条黑狗~(A man with a camera stood by a man holding a umbrella and a black dog) \\
 & AIC-ICC: 一个戴着帽子的男人和一个右手拿着伞的女人走在道路上~(A man in a hat and a woman with an umbrella in her right hand walk on the road)\\
 & COCO-MT: 一个男人和一个女人拿着一把伞~(A man and a woman hold an umbrella)\\
 & COCO-CN: 两个男人和一个女人坐在长椅上~(Two men and a woman sitting on a bench)\\
 & COCO-Mixed: 一个 男人 和 一个 女人 站 在 一 把 伞 下 \\
 & Sequential Learning: 两个男人和一只狗站在一把伞下~(Two men and a dog stand under an umbrella)\\

\cmidrule{2-2}

 \multirow{10}{*}{\includegraphics[width=3cm]{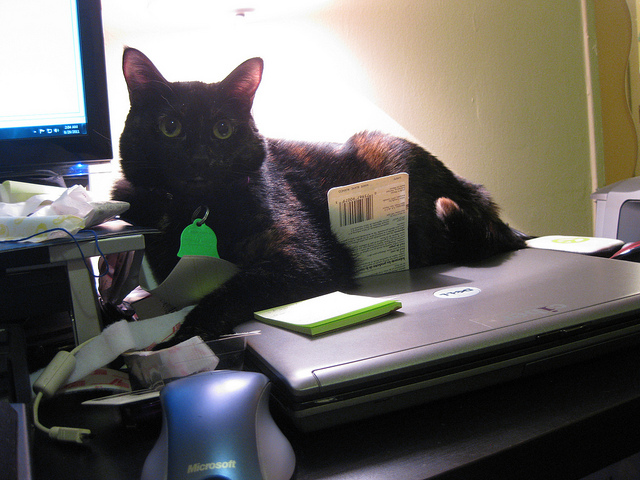}} 
 & \emph{\textbf{Image Tagging}}: \\ 
 & Manual: 猫 (cat), 笔记本电脑 (laptop)\\
 & COCO-MT: 猫 (cat), 电脑 (computer), 笔记本 (notebook), 桌子 (table), 键盘 (keyboard) \\
 & COCO-CN: 电脑 (computer), 猫 (cat), 笔记本 (notebook), 笔记本电脑 (laptop), 书桌 (desk) \\
 & Clarifai: 计算机 (computer), 笔记本电脑 (laptop), 商业 (business), 技术 (technology), 房间 (room) \\
 & Multi-task MLP: 猫 (cat), 电脑 (computer), 笔记本 (notebook), 笔记本电脑 (laptop), 书桌 (desk) \\
 & Cascading MLP: 电脑 (computer), 猫 (cat), 笔记本 (notebook), 笔记本电脑 (laptop),  书桌 (desk) \\ [1pt]
 & \emph{\textbf{Image Captioning}}: \\
 & Manual: 一只黑猫趴在一台笔记本电脑上~(A black cat is lying on a laptop) \\
 & AIC-ICC: 房间里有一个坐在椅子上的女人在看电脑~(There is a woman sitting on the chair in the room watching the computer)\\
 & COCO-MT: 躺在笔记本电脑上的猫~(The cat lying on the laptop)\\
 & COCO-CN: 一只猫趴在笔记本电脑上~(A cat is lying on the laptop)\\
 & COCO-Mixed: 一只黑猫坐在笔记本电脑旁边的桌子上~(A black cat sits on the table next to the laptop) \\
 & Sequential Learning: 一只黑猫趴在笔记本电脑上~(A black cat is lying on the laptop)\\

\cmidrule{2-2}

 \multirow{10}{*}{\includegraphics[width=3cm]{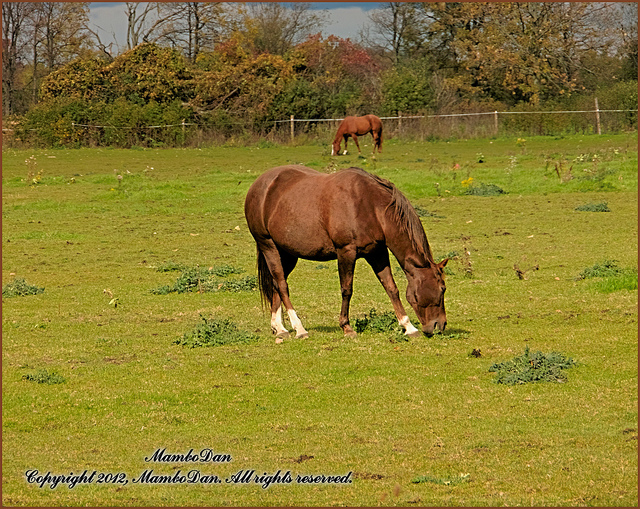}} 
 & \emph{\textbf{Image Tagging}}: \\ 
 & Manual: 马 (horse), 棕色 (brown), 草场 (meadow) \\
 & COCO-MT: 马(horse), 草地 (grassland), 田野 (field), 棕色(brown), 草 (grass)\\
 & COCO-CN: 马(horse), 草地 (grassland), 草 (grass), 棕色 (brown), 草场 (meadow) \\
 & Clarifai: 放牧 (graze), 草坪 (lawn) 哺乳动物 (mammal), 沒有人 (no person), 乡村的 (countryside) \\
 & Multi-task MLP: 马(horse), 草地 (grassland), 草 (grass), 棕色 (brown), 草场 (meadow) \\
 & Cascading MLP: 马(horse), 草 (grass), 草地 (grassland), 棕色 (brown), 草场 (meadow)\\ [1pt]
 & \emph{\textbf{Image Captioning}}: \\
 & Manual: 两只棕色的马在草地上吃草~(Two brown horses graze on the grassland) \\
 & AIC-ICC: 绿油油的草地上有一个穿着白色上衣的女人在骑马~(On the green grass there is a woman in a white coat riding a horse)\\
 & COCO-MT: 在草地上吃草的马~(A horse that grazes on the grassland)\\
 & COCO-CN: 一匹马站在草地上~(A horse stands on the grassland)\\
 & COCO-Mixed: 一匹棕色的马站在一片绿油油的田野上~(A brown horse stands on a green field) \\
 & Sequential Learning: 一匹棕色的马在草地上吃草~(A brown horse grazes on the grassland)\\

\cmidrule{2-2}

 \multirow{10}{*}{\includegraphics[width=3cm]{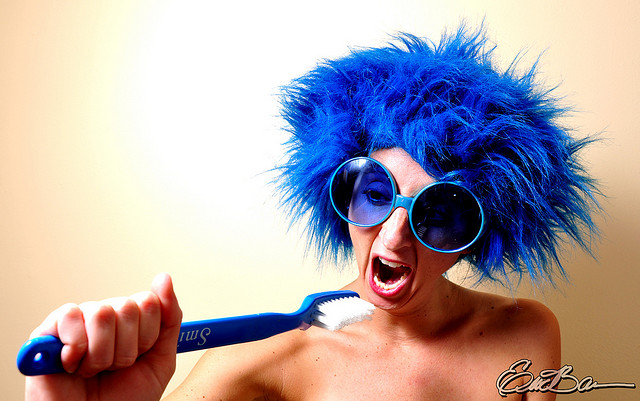}} 
 & \emph{\textbf{Image Tagging}}: \\ 
 & Manual: 墨镜 (sunglass), 蓝色 (blue), 头发 (hair), 牙刷 (toothbrush), 女孩 (girl) \\
 & COCO-MT: 牙刷 (toothbrush), 刷牙 (brush one's teeth), 蓝色 (blue), 女人 (woman), 眼镜 (glass) \\
 & COCO-CN: 牙刷 (toothbrush), 刷牙 (brush one's teeth), 蓝色 (blue), 剪刀 (scissor), 墨镜 (sunglass) \\
 & Clarifai: 时尚 (fashion), 太阳镜 (sunglass), 乐趣 (fun), 女人 (woman),  肖像 (portrait) \\
 & Multi-task MLP: 牙刷 (toothbrush), 刷牙 (brush one's teeth), 蓝色 (blue), 女人 (woman), 剪刀 (scissor) \\
 & Cascading MLP: 牙刷 (toothbrush), 蓝色 (blue), 刷牙 (brush one's teeth), 女人 (womaon), 头发 (hair) \\ [1pt]
 & \emph{\textbf{Image Captioning}}: \\
 & Manual: 一位蓝色头发带着蓝色墨镜的女孩想用巨大的蓝色牙刷刷牙~(A girl with blue hair and blue sunglasses wants to brush her teeth with a huge blue toothbrush) \\
 & AIC-ICC: 房间里有一个戴着墨镜的女人在喝水~(There is a woman wearing sunglasses in the room drinking water)\\
 & COCO-MT: 一个戴着牙刷的人~(A person wearing a toothbrush)\\
 & COCO-CN: 一只手拿着一把牙刷~(A toothbrush in one hand)\\
 & COCO-Mixed: 一个戴着眼镜的女人在刷牙~(A woman with glasses is brushing her teeth) \\
 & Sequential Learning: 一个穿着蓝色衣服的女人正在刷牙~(A woman in blue is brushing her teeth)\\

\cmidrule{2-2}

 \multirow{10}{*}{\includegraphics[width=3cm]{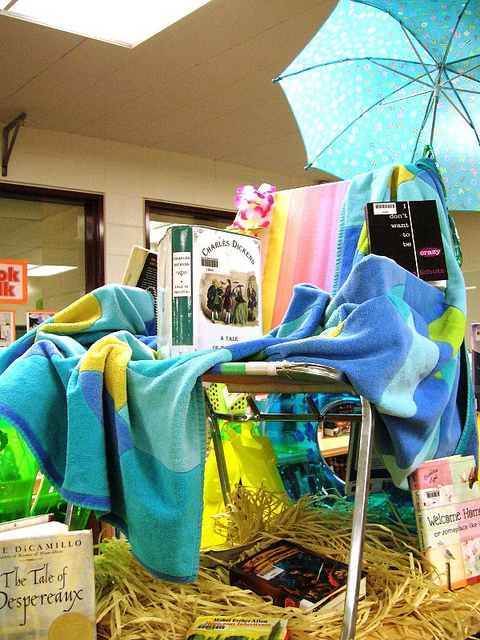}} 
 & \emph{\textbf{Image Tagging}}: \\ 
 & Manual: 雨伞 (umbrella), 布料 (cloth), 房子 (house), 草 (grass) \\
 & COCO-MT: 毛巾 (towel), 雨伞 (umbrella), 桌子 (table), 黄色 (yellow), 蓝色 (blue) \\
 & COCO-CN: 蓝色 (blue), 椅子 (chair), 玩具 (toy), 雨伞 (umbrella), 伞 (umbrella) \\
 & Clarifai: 沒有人 (no person), 购物 (shopping), 市场 (market), 存储 (storage), 廉价出售 (undersell) \\
 & Multi-task MLP: 椅子 (chair), 玩具 (toy), 蓝色 (blue), 伞 (umbrella), 玩偶 (doll) \\
 & Cascading MLP: 雨伞 (umbrella), 蓝色 (blue), 椅子 (chair), 桌子 (table), 摆放 (display) \\ [1pt]
 & \emph{\textbf{Image Captioning}}: \\
 & Manual: 一间房子里有一把伞，椅子上有一些布料，地上有一些草~(There is an umbrella in a house, with some cloth on the chair and some grass on the ground.) \\
 & AIC-ICC: 房间里有一个穿着蓝色衣服的女人在看书~(There is a woman in blue in the room reading)\\
 & COCO-MT: 有许多不同颜色的雨伞的房间~(Rooms with umbrellas of many colors)\\
 & COCO-CN: 房间里放着许多毛绒玩具~(There are many stuffed toys in the room)\\
 & COCO-Mixed: 有许多不同颜色的雨伞的房间~(Rooms with umbrellas of many colors)\\
 & Sequential Learning: 房间里有一把蓝色的雨伞和一把椅子~(There is a blue umbrella and a chair in the room)\\

 \bottomrule
\end{tabular}
}
\end{table*}

\subsection{Task II: Cross-Lingual Image Captioning} \label{ssect:cap}
Naturally, COCO-CN is suitable for research on image captioning in Chinese and in a cross-lingual setting. To demonstrate the necessity of COCO-CN, we present the following experiments, comparing models trained on different datasets. 

\subsubsection{Setup}

\textbf{Test set}. Each of the 1,000 test images in COCO-CN is associated with one manually written Chinese sentence at least. To enrich the ground truth, we manually translated the existing five English sentences per test image. Consequently, each test image has six Chinese sentences as ground truth. 

\textbf{Captioning model}. We adopt the Show and Tell network \cite{google-show-tell}, originally developed for English caption generation and later found to be effective for Chinese caption generation~\cite{flickr8kcn,weiyu2017fluency,wu2017ai}. The network computes the posterior probability of a sentence $S$ given an input image $x$ by combining a CNN based image encoder and an LSTM based sentence decoder. Consequently, the image \blue{is} annotated with the sentence \blue{yielding} the maximal probability. 
The size of LSTM, \ie the dimensionality of its hidden vector, is empirically set to 512.

\textbf{Training strategies}. As mentioned above, Flickr8k-cn~\cite{flickr8kcn} and AIC-ICC~\cite{wu2017ai} already exist for image captioning in Chinese. One more alternate is COCO-MT, a machine translation version of MS-COCO. We compare COCO-CN with these three datasets. For a fair comparison the Show and Tell model is separately trained using each dataset. The model is trained  using standard supervised learning that minimizes the cross entropy loss.  The initial learning rate is set to be 0.0005, which decays every three epochs with a decaying factor of 0.8.

While the machine-translated sentences in COCO-MT tend to contain grammatical errors and do not read naturally, this dataset has a much larger vocabulary (with 10,144 words occurring at least 5 times) for training. In order to effectively exploit the large-scale property of COCO-MT and the high-quality property of COCO-CN, we introduce a \emph{Sequential Learning} strategy. Using the joint vocabulary of the two datasets, we first train the captioning model using COCO-MT. After 30 epochs, we pick up the model that has the highest CIDEr score on the COCO-CN validation set. We continue to train this model using COCO-CN, with a learning rate of 0.00005. Sequential Learning is in some way similar to fine tuning,  with its focus on appropriate utilization of machine-translated and manually written training sentences. From this viewpoint, Sequential Learning is not \red{new by itself}. Rather, it provides a novel and more effective way to exploit cross-lingual resources, when compared to existing works \cite{miyazaki2016cross,multilingual-caption}.

\textbf{Baselines}. To the best of our knowledge, the \emph {Transfer learning} model \cite{miyazaki2016cross} and the \emph{Artificial token} model \cite{multilingual-caption} are the only two existing alternatives that target at learning from cross-lingual resources, both originally proposed for generating Japanese captions. In our implementation, the former initializes the visual embedding matrix of a Chinese captioning model using its counterpart in a trained English captioning model, while the latter learns a bilingual model from both manually written Chinese captions and original MS-COCO English captions. In addition, one might consider \red{having} COCO-MT and COCO-CN simply blended and used for training the image captioning model in one stage. We term this strategy COCO-Mixed, a necessary baseline to investigate whether the sequential learning strategy really matters.

\textbf{Evaluation criteria}. We use standard automatic evaluation metrics to evaluate caption quality, namely BLEU, METEOR, ROUGE-L and CIDEr. While being convenient, the automated metrics have weak correlation\red{s} with human judgments of caption quality. We perform a user study, in terms of relevance and fluency, on COCO-CN test100 and NUS-WIDE100. Fluency reflects the extent to which a sentence reads naturally. The same eight subjects participated \red{in} the user study. Each sentence was independently rated by two subjects on a Likert scale of 1 to 5 (higher is better) for relevance and fluency.  To help a subject give more comparable scores, each test image and sentences generated by different models were presented together to the subject. To avoid bias, the sentences were randomly shuffled in advance.

\subsubsection{Results}

\textbf{Automated evaluation}. Table \ref{tab:cap_autoeval} shows automated evaluation results of different captioning models.
Models trained on Flickr8k-CN and AIC-ICC have fairly low scores.
The \red{relatively} limited size of Flickr8k-CN makes it lack the ability for training a good captioning model. As for AIC-ICC, although it is the largest Chinese captioning dataset, it is strongly biased that all the images are about human beings. Consequently, the model trained on this dataset is unsuitable for describing general images. 
The model trained on COCO-CN is on par with the model trained on machine translated sentences of the full MS-COCO set.
As for COCO-Mixed, since manually written sentences from COCO-CN are overwhelmed by machine translated sentences from COCO-MT during training, the benefit of COCO-CN appears to be marginal. By contrast, Sequential learning is quite effective, performing best under all metrics.

\textbf{Human evaluation}. Table \ref{tab:cap_humaneval} shows human evaluation of distinct models on the two test sets. The COCO-MT model receives the lowest fluency score of 4.50 on COCO-CN test100 and 4.36 on NUS-WIDE100. 
Errors in machine translation impair the accuracy and grammar of the sentences generated by the COCO-MT model. With sentences carefully written by native Chinese speakers and covering a wide variety of visual content, COCO-CN is suitable for generic Chinese image captioning. Sequential learning allows us to effectively leverage the manual annotations. While its fluency score of 4.76 is lower than the COCO-CN counterpart on COCO-CN test100 and that of AIC-ICC on NUS-WIDE100, sequential learning gives the best relevance score in terms of both automatic and human evaluation, suggesting that the generated captions are the most descriptives.

\begin{table} [tb!]
\renewcommand{\arraystretch}{1.1}
\caption{\textbf{Automated evaluation of image captioning models trained on different datasets}. The proposed cross-lingual transfer performs the best.}
\label{tab:cap_autoeval}
\centering
 \scalebox{0.9}{
\begin{tabular}{@{}l r r r r @{}}
\toprule
\textbf{Training}     & \textbf{BLEU-4} & \textbf{METEOR} & \textbf{ROUGE-L} & \textbf{CIDEr}\\
\midrule
Flickr8k-CN  & 10.1  & 14.9   & 33.8  & 22.9  \\
AIC-ICC      &  7.4  & 21.3   & 34.2  & 24.6  \\
COCO-MT      & 30.2  & 27.1   & 50.0  & 86.2  \\ 
COCO-CN      & 31.7  & 27.2   & 52.0  & 84.6  \\
COCO-Mixed   & 29.8  & 28.6   & 50.3  & 86.8  \\
Transfer learning \cite{miyazaki2016cross}  & 33.7 & 28.2 & 52.9 & 89.2 \\
Artificial token \cite{multilingual-caption} & 31.8 &  26.9 &  51.5 & 85.4 \\ 
\emph{Sequential Learning} & \textbf{36.7} & \textbf{29.5}	& \textbf{55.0} &	\textbf{98.4} \\
\bottomrule
\end{tabular}
 }
\end{table}

\begin{table} [tb!]
\renewcommand{\arraystretch}{1.1}
\caption{\textbf{Human evaluation of image captioning models on two test sets}. The numbers after $\pm$ are standard deviations. Sequential learning obtains the best overall score.}
\label{tab:cap_humaneval}
\centering
 \scalebox{0.9}{
\begin{tabular}{@{}l r r r r@{}}
\toprule
                    & \multicolumn{2}{c}{COCO-CN test100} & \multicolumn{2}{c}{NUS-WIDE100} \\
                    \cmidrule{2-3}  \cmidrule{4-5}
Model               &  Relevance         & Fluency &  Relevance         & Fluency\\
\midrule
Flickr8k-CN         & 2.34 $\pm 1.10$    & 4.62 $\pm 0.62$        & 1.96 $\pm 1.03$     & 4.71 $\pm 0.48$ \\
AIC-ICC             & 2.63 $\pm 1.24$    & 4.75 $\pm 0.44$        & 2.17 $\pm 1.10$     & \textbf{4.73} $\pm 0.55$ \\
COCO-MT             & 3.89 $\pm 0.77$    & 4.50 $\pm 0.77$        & 2.92 $\pm 1.18$     & 4.36 $\pm 0.86$ \\
COCO-CN             & 3.95 $\pm 0.90$  & \textbf{4.83} $\pm 0.51$ & 2.69 $\pm 1.08$     & 4.59 $\pm 0.72$ \\
Transfer Learning \cite{miyazaki2016cross}      & 3.89 $\pm 0.83$   & 4.78 $\pm 0.49$    & 2.86 $\pm 1.10$     & 4.62 $\pm 0.61$ \\
Artificial token \cite{multilingual-caption}    & 3.78 $\pm 0.99$   & 4.78 $\pm 0.69$    & 2.61 $\pm 1.10$     & 4.46 $\pm 0.90$ \\
\emph{Sequential Learning} & \textbf{4.13} $\pm 0.72$  & 4.76 $\pm 0.59$ & \textbf{3.02 $\pm 1.15$}     & 4.67 $\pm 0.60$ \\
\bottomrule
\end{tabular}
 }
\end{table}

Examples of captions generated by different models are shown in Table \ref{tab:tagging-example}. Consider the second row for instance. The manually written sentence is 一只黑猫趴在一台笔记本电脑上~(A black cat is lying on a laptop). The AIC-ICC model predicts 房间里有一个坐在椅子上的女人在看电脑~(There is a woman sitting on the chair in the room watching the computer), showing a strong bias of the model on describing human actions. Compared to the COCO-CN model which describes the image as 一只猫趴在笔记本电脑上~(A cat is lying on the laptop), both COCO-Mixed and Sequential Learning is more descriptive by predicting 黑猫~(black cat). The result shows the benefit of learning from the cross-lingual resources. Moreover, compared to COCO-Mixed which uses the verb 坐~(sit), Sequential Learning uses the verb 趴~(lying), which is more precise and vivid in the Chinese language.

\subsection{Task III: Cross-Lingual Image Retrieval} \label{ssect:retrieval}

\subsubsection{Setup}

The bilingual property of COCO-CN supports cross-lingual image retrieval. We consider in this paper the following setting. Given a Chinese sentence as a query, the goal is to find amidst a set of images the one best matches the query. Each of the images is described by \red{an} English sentence. \blue{The proposed cross-lingual image retrieval differs from a monolingual task as the former has to look into not only similarities between the visual and textual modalities but also similarities between texts presented in two distinct languages. To accomplish the proposed task, one needs to effectively represent Chinese queries, images and their English descriptions in a common space by cross-modal, cross-lingual representation learning. The task is thus of more research interest than its monolingual counterpart}. While cross-modal matching is well known to be difficult, image retrieval by cross-lingual matching is also challenging. Even if the Chinese and English sentences are meant for describing the same image, they were independently written by distinct subjects, presumably with varied educational and cultural background. So they are not necessarily semantically aligned. A desirable method shall exploit cross-modal matching between the query and a given image and cross-lingual matching between the query and the English description of the given image. 

\textbf{Test set}. We again use the test set of COCO-CN. For each image, its manually written Chinese sentence is taken as a query, resulting in a query set of size 1,000. Recall that each image has five English sentences from MS-COCO. We take the first sentence as the English description of the image.

\textbf{Retrieval models}. We see three approaches. One, cross-lingual (CL) matching between the Chinese query and the English description, with the visual content ignored. Two, cross-modal (CM) matching between the query and the image, with the English description ignored. Third, cross-lingual and -modal (CLM) matching that combines the previous two approaches. W2VV previously used for sentence recommendation can project a given sentence into a visual feature space. Having sentences of distinct languages projected into the same visual feature space enables both cross-lingual and cross-modal matching. Hence, we adopt W2VV as a unified solution. 

\begin{figure}[tb!]
    \centering
    \includegraphics[width=\columnwidth]{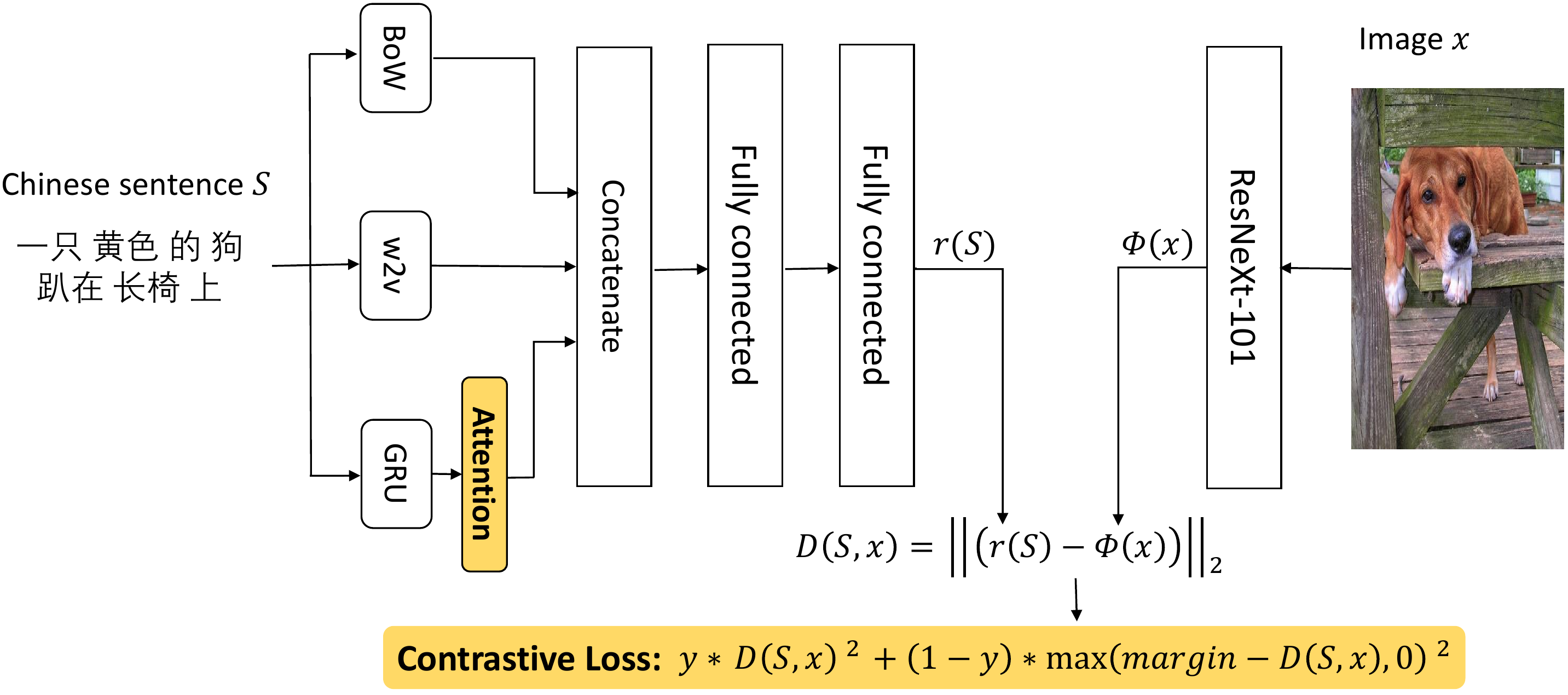}
    \caption{\textbf{W2VV and our improvement}. We enhance W2VV ~\cite{w2vv} by 1) adding an attention layer after GRU to take into account all the hidden vectors, and 2) substituting the original mean square error for the contrastive loss. The enhanced W2VV is used for cross-lingual image retrieval.}
    \label{fig:w2vv}
\end{figure}

We start with the best configuration of W2VV~\cite{w2vv}. An input sentence is vectorized in parallel by three strategies: bag-of-words (BoW), Word2Vec (w2v) and a Gated Recurrent Units (GRU) network, as Fig. \ref{fig:w2vv} shows. For GRU based vectorization, W2VV uses the hidden vector at the last time step, which is now known to be over compact to represent the entire sentence \cite{attention2015,ijcnlp17Elliott}. 
Consider a query sentence ``三个塑料饭盒中装着面条，蔬菜和水果''~(Three plastic lunch boxes containing noodles, vegetables and fruits) for instance. While the query
requires co-presence of multiple objects such as 塑料饭盒~(plastic lunch boxes), 面条~(noodles), 蔬菜~(vegetables) and 水果~(fruits), W2VV tends to over-emphasize the ending part of the query, \ie fruits.
Moreover, W2VV is trained using the Mean Squared Error (MSE), which does not consider negative pairs, \ie images and sentences \red{that} are irrelevant. \blue{Based on these two observations}, we enhance the \blue{encoding module and the loss} of W2VV as follows. 

First, we improve its GRU part by introducing a soft attention mechanism \cite{attention2015}. As shown in Fig \ref{fig:w2vv}, the attention layer is placed right after GRU, summing up the hidden vectors over all time steps with an adaptive weight assigned to each vector. More formally, let $h_i$ be the hidden vector at time step $i$, with $i=1,\ldots,n$ and $n$ is the maximum time step. The corresponding weight $\alpha_i$ is computed as 
\begin{equation}
\begin{array}{l}
\displaystyle m_i  = W^T h_i + b,\\
\displaystyle [\alpha_1, \alpha_2, \ldots, \alpha_n] = softmax([m_1,m_2,\ldots,m_n]),
\end{array} 
\label{eq:attention}
\end{equation}
where $W$ and $b$ parameterize the affine transformation within the attention layer. Accordingly we substitute $h_n$ in W2VV for $\sum_{i=1}^n \alpha_i h_i$.

Second, in order to exploit the many negative pairs, we substitute the MSE loss for the Contrastive Loss \cite{cvpr06-contrastive-loss}. Given an image $x$ and a sentence $S$, we use a binary variable $y$ to indicate their relevance, i.e., $y=1$ if they are relevant and $0$ otherwise. The new loss is
\begin{equation} \label{eq:contrastive-loss}
y \cdot D(S,x)^2 + (1-y) \cdot \max(margin-D(S,x), 0)^2,
\end{equation} 
where $D(S,x)$ indicate the Euclidean distance between the image and sentence vectors and $margin$ is a hyper-parameter, empirically set to be 2. 
Minimizing Eq. \ref{eq:contrastive-loss} means to minimize (maximize) the distance between relevant (negative) pairs.

We use COCO-CN to train our Chinese model that transforms a given Chinese sentence into a \resnext~feature. In a similar manner, an English version of our model is trained using the images from the same training set but with MS-COCO sentences. The cross-lingual (CL) similarity between a Chinese sentence and an English sentence is computed as the cosine similarity between their \resnext~features. Similarly, the cross-modal (CM) similarity between a Chinese sentence and an image is computed.

Given a Chinese query, image retrieval by cross-lingual matching is achieved by sorting the images in descending order according to the CL similarities of their English sentences to the query. Then, image retrieval by cross-modal matching is to sort images in descending order according to their CM similarities to the query. Finally, for image retrieval by cross-lingual and -modal (CLM) matching, the similarity between the query and a given image is computed as a convex combination of the CL and CM similarities. The combination weight is optimized on the validation set.

\textbf{Evaluation criterion}. Per query there is only one image known to be relevant with respect to the query. Hence, we report Mean Reciprocal Rank (MRR) at which the relevant item is found. Higher MRR means better performance. 

\subsubsection{Results}

Table \ref{tab:retrieval} \red{shows} performance of the different approaches. The Enhanced W2VV outperforms the original W2VV, justifying the effectiveness of the joint use of the attention mechanism and the contrastive loss in the new context. Image retrieval by CM is less effective than its CL counterpart. The best performance is achieved by CLM matching, scoring the highest MRR value of 0.511. This is a relative improvement of 6.2\% over the original W2VV. Some qualitative results are given in Table \ref{tab:image-rank}. The Enhanced W2VV better models relatively complex queries that require co-occurrence of multiple visual concepts. Notice the exception in the second row, where the ending part of the query, \ie 红色环~(red ring), is more important. In this case, the shortcoming of W2VV of over-emphasizing the ending part helps.

\begin{table} [tb!]
\renewcommand{\arraystretch}{1.1}
\caption{\textbf{Performance of different approaches to cross-lingual image retrieval}, measured by MRR.}
\label{tab:retrieval}
\centering
 \scalebox{0.95}{
\begin{tabular}{@{}l rrr @{}}
\toprule
         & \multicolumn{3}{c}{\textbf{Matching approaches}} \\
         \cmidrule{2-4}
\textbf{Model} & \emph{CL} & \emph{CM} & \emph{CLM} \\  
 \cmidrule{1-1}  \cmidrule{2-4}
W2VV~\cite{w2vv}      & 0.450 & 0.378  & 0.481 \\
Enhanced W2VV (\emph{attention}) & 0.466 &	0.412 & 0.499 \\
Enhanced W2VV (\emph{attention} + \emph{contrastive loss}) & 0.468 & 0.421  & \textbf{0.511} \\ 
\bottomrule
\end{tabular}
 }
\end{table}

\begin{table}[tb!]
\renewcommand{\arraystretch}{1.1}
\centering
\caption{\textbf{Enhanced W2VV versus original W2VV for image retrieval by cross-lingual and -modal (CLM) matching}. Texts in parentheses are English translations, provided for non-Chinese readers and not used for image retrieval. While W2VV tends to over-emphasize the ending words of a query sentence, Enhanced W2VV with the attention mechanism better models the entire query.}
\label{tab:image-rank}
\scalebox{0.8}{
\begin{tabular}{@{}lll@{}}
\toprule
\textbf{Chinese sentence as query} & \multicolumn{2}{c}{\textbf{Top hit}} \\ 
\cmidrule{2-3}
               & \emph{by W2VV} & \emph{by Enhanced W2VV} \\
\cmidrule{1-3}           
\multirow{1}{0.5\columnwidth}{\specialcell{三个 塑料 饭盒 中装 着 面条 ，\\ 蔬菜 和 水果~(Three plastic lunch \\ boxes containing noodles, vegetables \\ and fruits)}} & \multirow{1}{*}{\includegraphics[width=2.4cm,height=2.4cm]{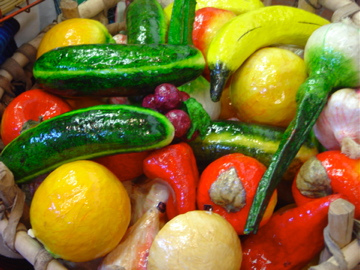}} & \multirow{1}{*}{\includegraphics[width=2.4cm,height=2.4cm]{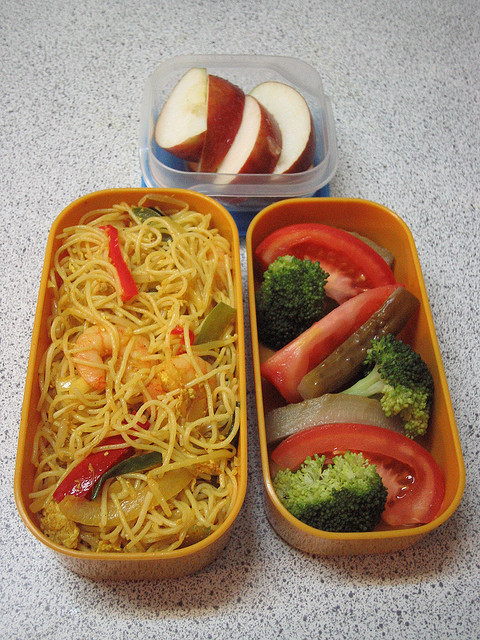}} \\ 
& \\
& \\
& \\
& \\
& \\
& \\

\multirow{1}{0.5\columnwidth}{\specialcell{停车场 标志牌 旁停 着  一辆 公交车\\~(Near the sign of  the parking lot \\ parked a bus)}} & \multirow{1}{*}{\includegraphics[width=2.4cm,height=2.4cm]{COCO_val2014_000000255769.jpg}} & \multirow{1}{*}{\includegraphics[width=2.4cm,height=2.4cm]{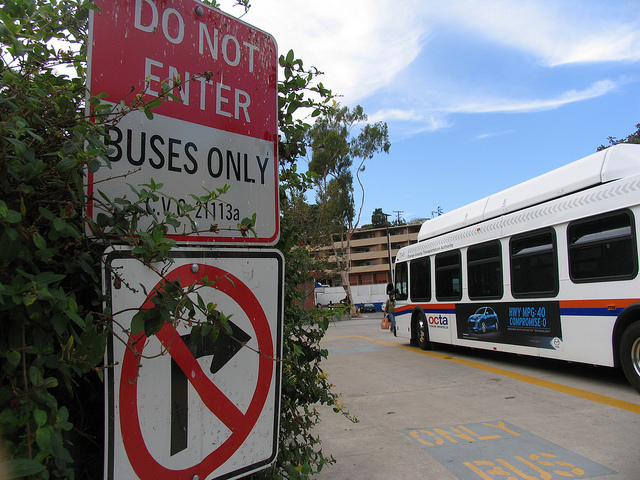}} \\ 
& \\
& \\
& \\
& \\
& \\
& \\

\multirow{1}{0.5\columnwidth}{\specialcell{一个 时钟 安装 在 一个 红色 环 中\\(A clock is installed in a red ring)}} & \multirow{1}{*}{\includegraphics[width=2.4cm,height=2.4cm]{COCO_train2014_000000402207.jpg}} & \multirow{1}{*}{\includegraphics[width=2.4cm,height=2.4cm]{COCO_train2014_000000523035.jpg}} \\ 
& \\
& \\
& \\
& \\
& \\
& \\

\multirow{1}{0.5\columnwidth}{\specialcell{城市 建筑 中间 的 一条 街道 上 ，竖立 着 \\ 一个 指示牌 ， 指示牌 下站 着 等 着  过 马路 \\ 的 男人 ~(On a street in the middle of  city \\ buildings, a sign is erected, under which \\ stood men waiting to cross the road)}} & \multirow{1}{*}{\includegraphics[width=2.4cm,height=2.4cm]{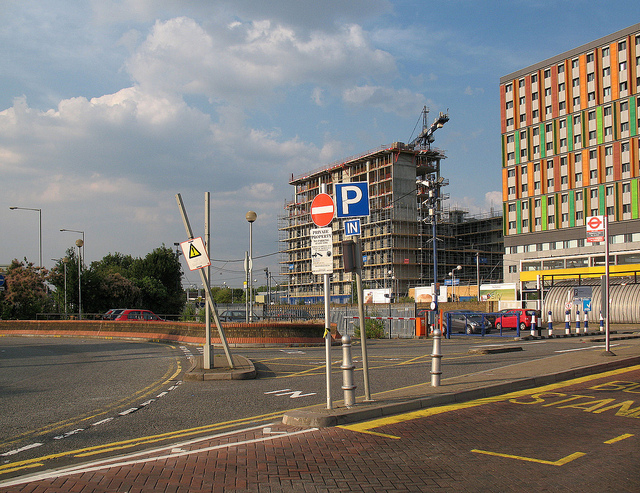}} & \multirow{1}{*}{\includegraphics[width=2.4cm,height=2.4cm]{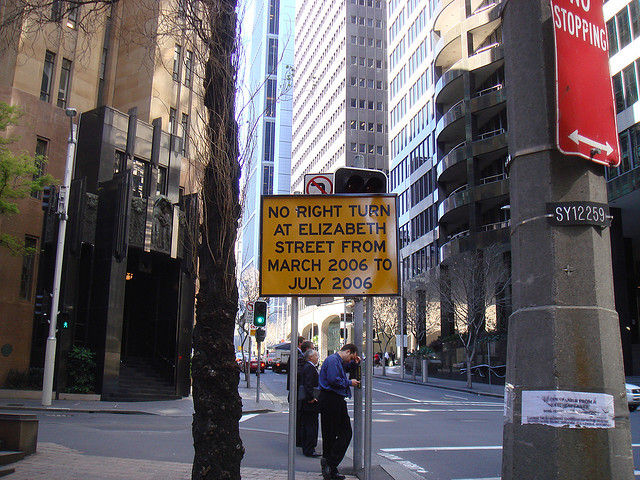}} \\ 
& \\
& \\
& \\
& \\
& \\ [8pt]

\bottomrule
\end{tabular}
}
\end{table}

\subsection{Discussion} \label{ssect:discuss}

To justify the viability of COCO-CN for cross-lingual image tagging, captioning and retrieval, we have developed baseline methods per task. While being conceptually simple, much engineering efforts are required to properly implement these methods. For reproducible research, we \blue{have released our source code at the same github URL as the dataset}.

There is much room for future exploration from an algorithmic perspective. We discuss a few recent methods related to image tagging, captioning and retrieval, albeit in a \emph{monolingual} setting. Wang \etal \cite{tomm14-bclda} propose a novel generative model, called Bilateral Correspondence Latent Dirichlet Allocation (BC-LDA), to find conditional relationships between images and texts based on good low-dimensional representations. BC-LDA is shown to be effective for microblog image tagging. In order to leverage partially labeled data for image and video annotation, Song \etal \cite{tip16-graph-learning} propose to learn an optimized graph from multiple cues. Relationships between data points are embedded into the graph more precisely. For action recognition in videos, Wang \etal \cite{tmm18-convnet-fusion} introduce two-stream 3-D ConvNet fusion that handles video clips with arbitrary size and length by spatial temporal pyramid techniques. To attack \red{the} data selection bias between the training and testing stages, Shen \etal \cite{mm18-data-bias} introduce causally regularized learning, which results in image classifiers with better generalization ability. In the context of video captioning, Gao \etal \cite{tmm17-gao} propose a novel attention-based LSTM model with semantic consistency to explore the correlation between multi-modal representations for generating sentences with rich semantic content. A more recent work by Song \etal \cite{tnnls18-video-captioning} improves video captioning by implementing the decoding module using multi-modal stochastic RNNs networks, \red{modeling} the uncertainty observed in the data using latent stochastic variables. For image retrieval, Cui \etal \cite{tois14-image-search} present an interesting study \red{on} how to transfer social knowledge sensed from social media platforms for reranking and consequently personalizing image search results from generic image search engines. For content-based video retrieval at large-scale, Song \etal \cite{tip18-video-hashing} propose self-supervised video hashing that simultaneously encodes video temporal and visual information using an end-to-end hierarchical binary auto-encoder and a neighborhood structure. For cross-modal retrieval, Peng \etal \cite{tmm18-ccl} propose to jointly employ coarse-grained instances and fine-grained patches within a multi-grained fusion network. Based on the observation that different modalities often contain \red{an} unequal amount of information when describing the same semantics, a modality-specific cross-modal similarity measurement approach is developed in \cite{tip18-cross-modal}, where an independent semantic space is constructed for each modality. The aforementioned algorithms have shown encouraging results for their own tasks, but all in a monolingual scenario. Repurposing and redesigning these algorithms to let them fit for the cross-ligual setting are beyond the scope of this paper.

Concerning the dataset size, we could have chosen crowd sourcing, which is actually more \red{cheaper} and more scalable, to make the numbers more impressive. However, as we noted in the paper and supported by our observations about MS-COCO annotations, crowd sourcing lacks quality control. We thus progressed with the proposed annotation approach, which produced high-quality annotations as demonstrated by \blue{the data analytics in Section \ref{ssec:data-analysis} and the multi-task experiments in this section}. Compared to the existing datasets, COCO-CN results in noticeably better models for multiple tasks.

\section{Conclusions \blue{and Perspectives}} \label{sect:conc}

The development of COCO-CN indicates that our recommendation-assisted annotation system helps reduce annotators' workload. In particular, 54 \red{percent} of the recommended tags were accepted by the annotators. Sentence recommendation is also helpful, albeit less effective than tag recommendation. 

COCO-CN has enriched MS-COCO, not just in terms of the language. Compared to the 80 MS-COCO object classes, there are 585 novel concepts. On average a COCO-CN sentence brings in 45.3\% keywords not covered by machine translation of MS-COCO sentences. While the number of its images seems to be small, its diverse visual content associated with high-quality annotations makes the dataset superior to Flickr8k-CN and AIC-ICC. \blue{Compared to NUS-WIDE, a popular multi-label dataset that has 81 labels and 2.4 labels per image, COCO-CN provides 655 labels with 4.4 labels per image. With the rich and (unaligned) bilingual tags and sentences, COCO-CN is unique to the community, providing a unified and challenging platform for advancing image annotation and retrieval in the cross-lingual direction}.

Even though the ground truth of the test set is incomplete, it is adequate for drawing qualitative conclusions about which model is better for cross-lingual image tagging. Cascading MLP outperforms MLP trained on the monolingual resources, multi-task MLP trained on bi-lingual resources, and Clarifai, a commercial image tagging \blue{service}. Our sequential learning strategy, \blue{while being a common practice and thus not new in the monolingual setting}, is the first solution that has successfully combined machine-translated and manual annotations for cross-lingual image captioning. The enhanced W2VV establishes a new baseline for cross-lingual image retrieval. COCO-CN, provided with these good baseline methods, opens up interesting avenues for future research on cross-lingual \blue{multimedia tasks}.

\vspace{3pt}
\textbf{Acknowledgments}.
The authors thank all six anonymous reviewers for their insightful comments and Miss Xinru Chen for performing typos check on COCO-CN sentences. 


\section*{APPENDIX}

\begin{table*}[tb!]
\renewcommand{\arraystretch}{1}
\centering
\caption{\textbf{Comparing three CNN features}. For a fair comparison, all the features are evaluated in the same setting, \ie MLP trained on COCO-CN for image tagging, the Show and Tell model trained on COCO-CN for image captioning, and the enhanced W2VV for image retrieval. Models using the ResNeXt-101 feature are the best.}
\label{tab:three-feats}
\scalebox{0.9}{
\begin{tabular}{@{}l rrr r rrrr r rrr@{}}
\toprule

\textbf{Image Feature} &  \multicolumn{3}{c}{\textbf{Image Tagging}} && \multicolumn{4}{c}{\textbf{Image Captioning}} &&  \multicolumn{3}{c}{\textbf{Image Retrieval}} \\
                      \cmidrule{2-4} \cmidrule{6-9} \cmidrule{11-13}
                     & \emph{Precision} & \emph{Recall} & \emph{F-measure} && \emph{Bleu-4}	& \emph{METEOR}	& \emph{ROUGE-L} &	\emph{CIDEr} && \emph{CL} &	\emph{CM}	& \emph{CLM} \\
                    \cmidrule{1-13}
\emph{GoogLeNet}     & 0.394 & 0.473 & 0.414 && 26.1 & 27.3	& 49.6 & 73.5 && 0.458 & 0.336 & 0.463 \\
\emph{ResNet-152}    & 0.407 & 0.496 & 0.430 && 26.5 & 27.3	& 49.8 & 74.4 && 0.450 & 0.362 & 0.479 \\
\emph{ResNeXt-101}   & \textbf{0.477} & \textbf{0.576} & \textbf{0.503} && \textbf{31.7} & \textbf{27.2} & \textbf{52.0} & \textbf{84.6} &&  \textbf{0.468} & \textbf{0.421} & \textbf{0.511} \\
\bottomrule
\end{tabular}
}
\end{table*}

\textbf{Choice of Image Features}. As shown in Table \ref{tab:three-feats}, for all the three tasks models using the ResNeXt-101 feature perform the best, when compared to their counterparts using either the GoogLeNet or ResNet-152 feature.

\textbf{Image Captioning with Tags}. As we have summarized in Table  \ref{tab:datasets}, a novel property of COCO-CN is that its images are annotated with tags, in addition to sentence descriptions. 
We conduct an extra experiment, termed image captioning with tags, to demonstrate that the tags can complement the image captioning task. The image captioning model previously trained on COCO-CN is re-used as a baseline. To construct a multi-modal input for a given image, we represent tags associated with this image by a bag-of-words (BoW) feature vector. Each dimension corresponds a specific tag, with its value set to one if the tag is present and zero otherwise. A multi-modal input is formed by concatenating the ResNeXt-101 feature and the BoW feature. As Table \ref{tab:cap-with-tags} shows, image captioning with tags performs better.

\begin{table} [tb!]
\renewcommand{\arraystretch}{1}
\caption{\textbf{Automated evaluation of image captioning with and without tags}. COCO-CN is used as training data. Including tags as input gives better performance.}
\label{tab:cap-with-tags}
\centering
 \scalebox{0.9}{
\begin{tabular}{@{}l r r r r @{}}
\toprule
\textbf{Input}     & \textbf{BLEU-4} & \textbf{METEOR} & \textbf{ROUGE-L} & \textbf{CIDEr}\\
\midrule
Image        & \textbf{31.7}  & 27.2   & 52.0  & 84.6  \\
Image + tags & 31.3  & \textbf{30.1}   & \textbf{53.2} & \textbf{90.0} \\
\bottomrule
\end{tabular}
 }
\end{table}

\bibliographystyle{IEEEtran}
\bibliography{IEEEabrv,cococn}


\begin{IEEEbiography}[{\includegraphics[width=1in,height=1.25in,clip,keepaspectratio]{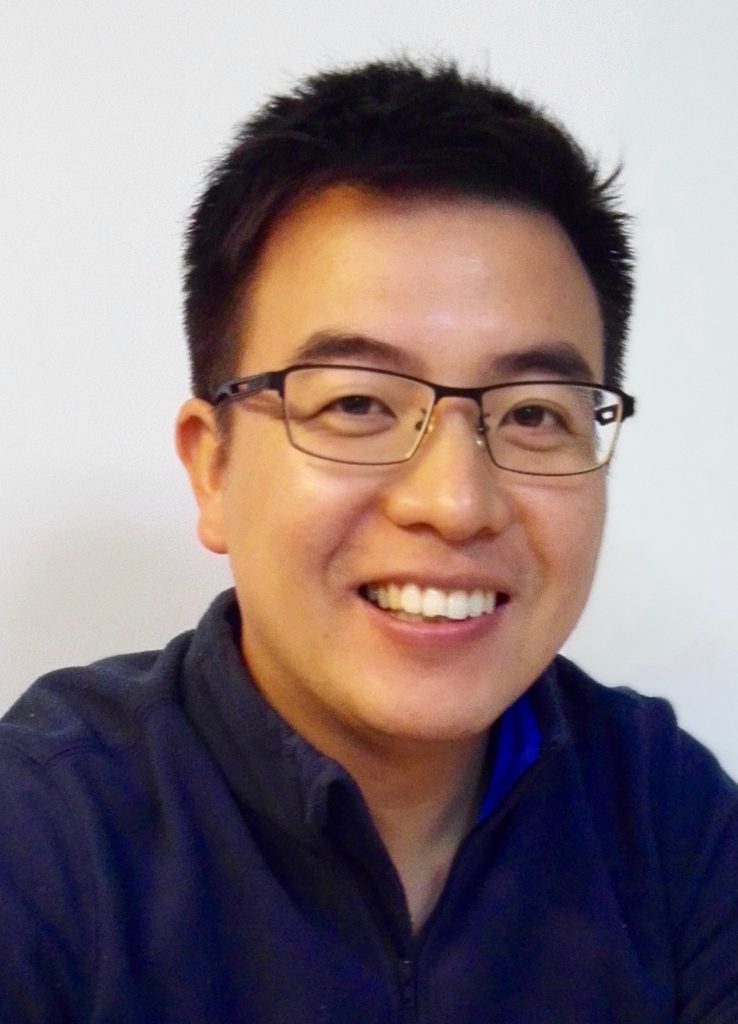}}]
{Xirong Li} received the B.S. and M.E. degrees from Tsinghua University, Beijing, China, in 2005 and 2007, respectively, and the Ph.D. degree from the University of Amsterdam, Amsterdam, The Netherlands, in 2012, all in computer science. He is currently an Associate Professor with the Key Lab of Data Engineering and Knowledge Engineering, Renmin University of China, Beijing, China. His research includes image and video retrieval.

Prof. Li was recipient of the ACMMM 2016 Grand Challenge Award, the ACM SIGMM Best Ph.D. Thesis Award 2013, the IEEE TRANSACTIONS ON MULTIMEDIA Prize Paper Award 2012, and the Best Paper Award of ACM CIVR 2010. He was area chair of ACMMM 2018 and ICPR 2016.

\end{IEEEbiography}

\begin{IEEEbiography}[{\includegraphics[width=1in,height=1.25in,clip,keepaspectratio]{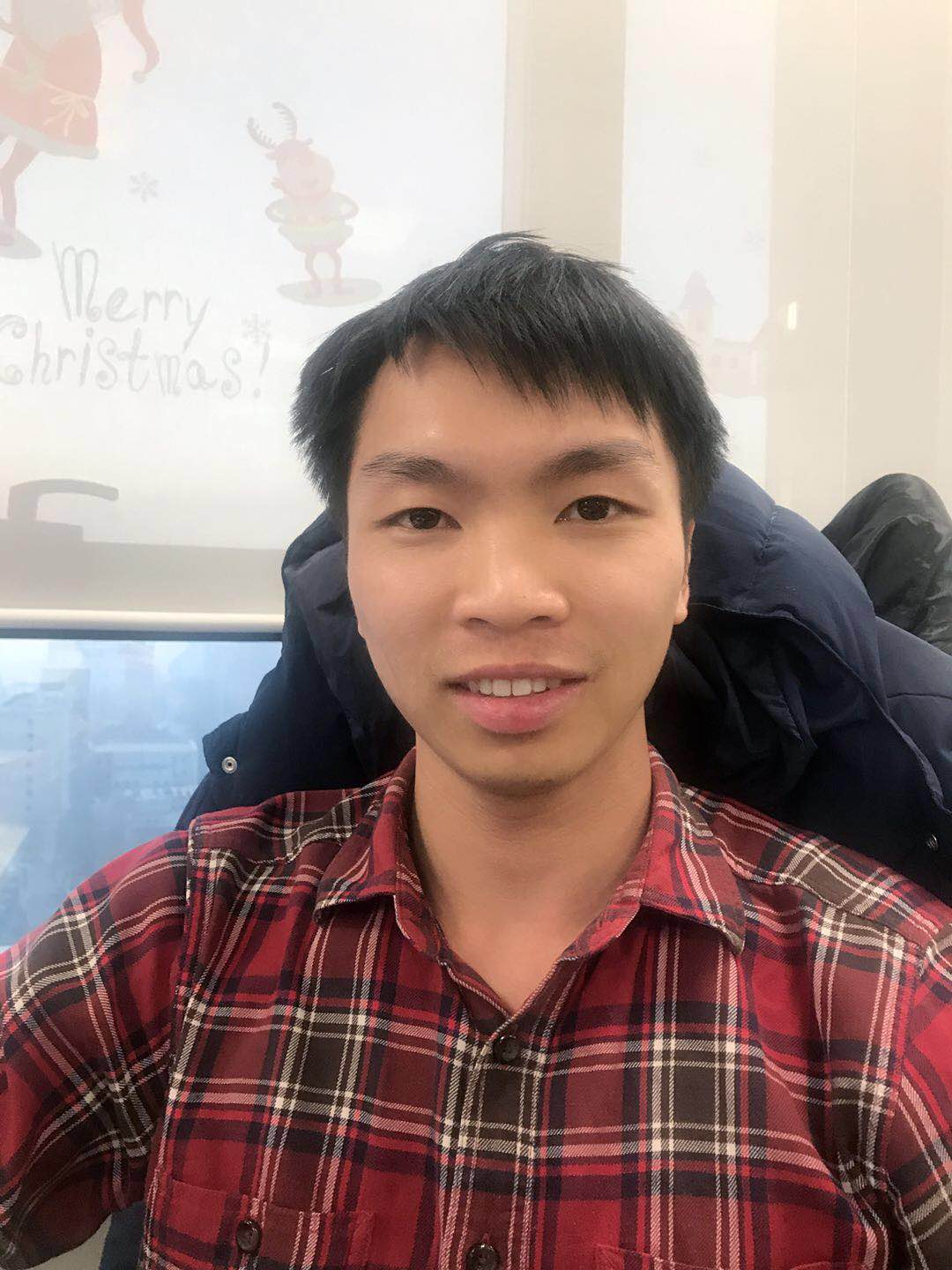}}]
{Chaoxi Xu} received his B.S. degree in Computer Science from Renmin University of China, Beijing, China in 2017. He is currently a graduate student at School of Information, Renmin University of China, pursuing his master degree on cross-lingual multimedia retrieval.
\end{IEEEbiography}

\begin{IEEEbiography}[{\includegraphics[width=1in,height=1.25in,clip,keepaspectratio]{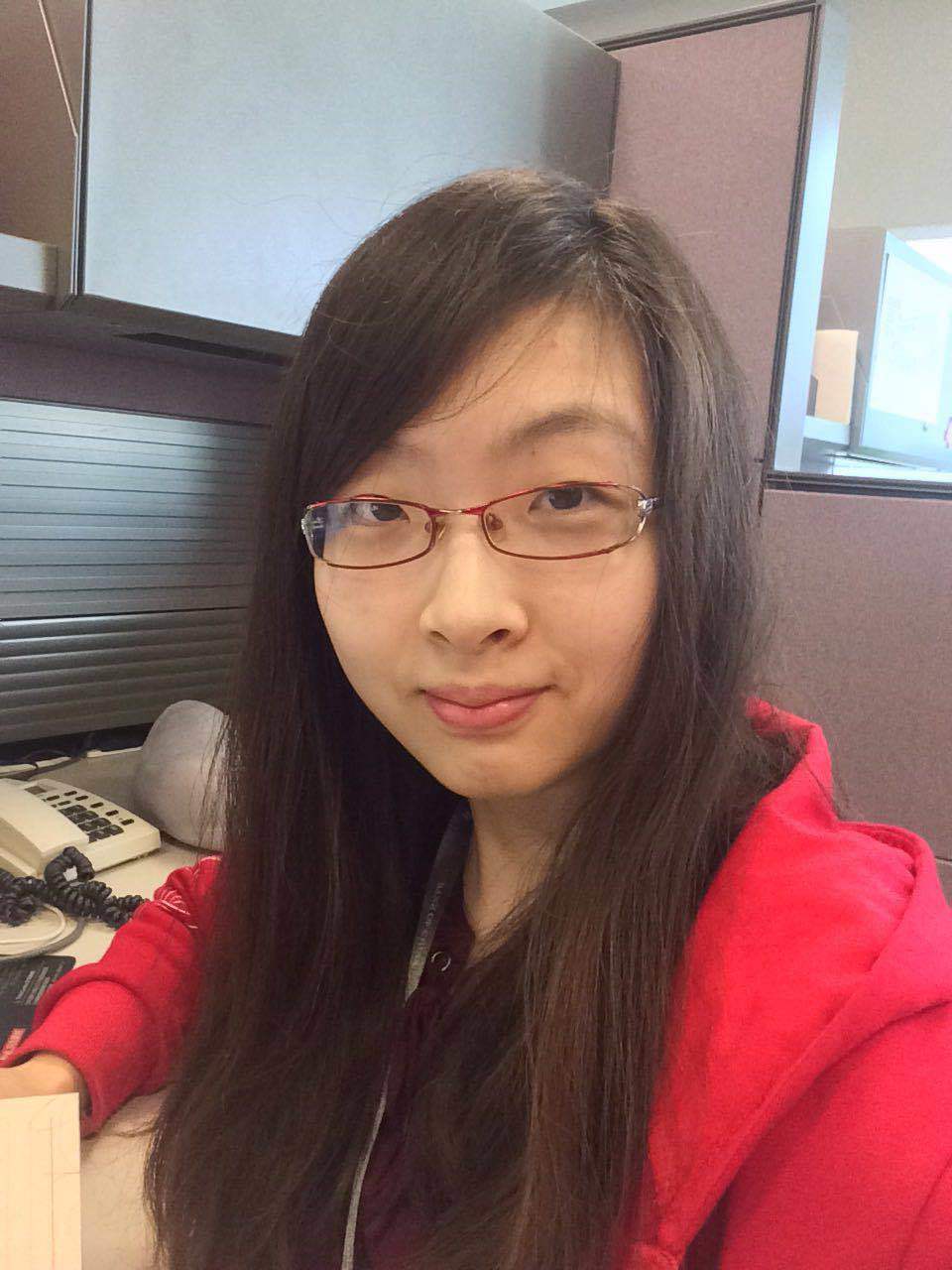}}]
{Xiaoxu Wang} received her B.S. and M.E. degrees in Computer Science from Renmin University of China, Beijing, China, in 2015 and 2018, respectively. She is currently a software engineer at the Bank of China, Beijing, China.
\end{IEEEbiography}

\begin{IEEEbiography}[{\includegraphics[width=1in,height=1.25in,clip,keepaspectratio]{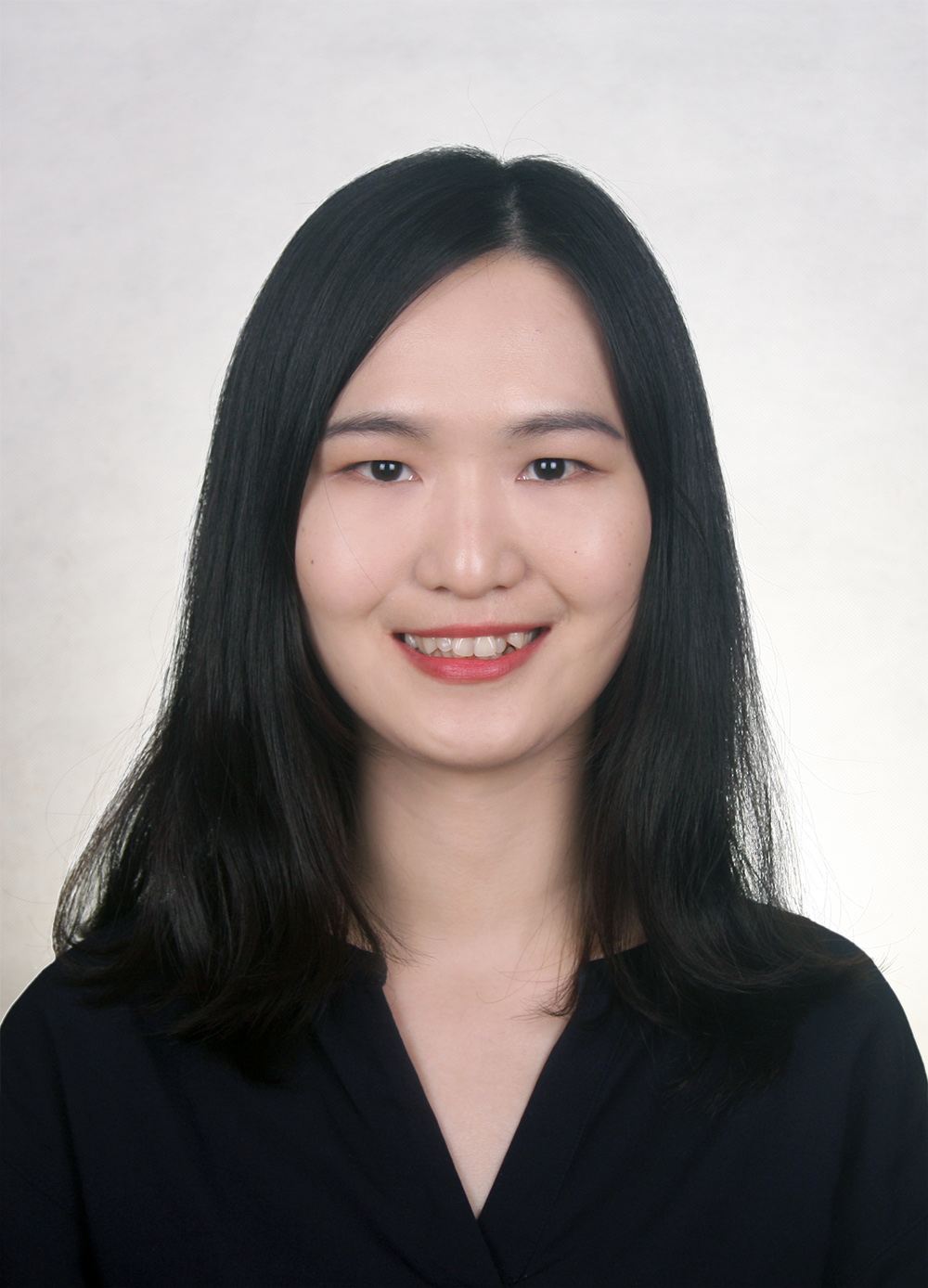}}]
{Weiyu Lan} received her B.S. and M.E. degrees in Computer Science from Renmin University of China, Beijing, China, in 2015 and 2018, respectively. She is currently a researcher at Tencent, Shenzhen, China.
\end{IEEEbiography}

\begin{IEEEbiography}[{\includegraphics[width=1in,height=1.25in,clip,keepaspectratio]{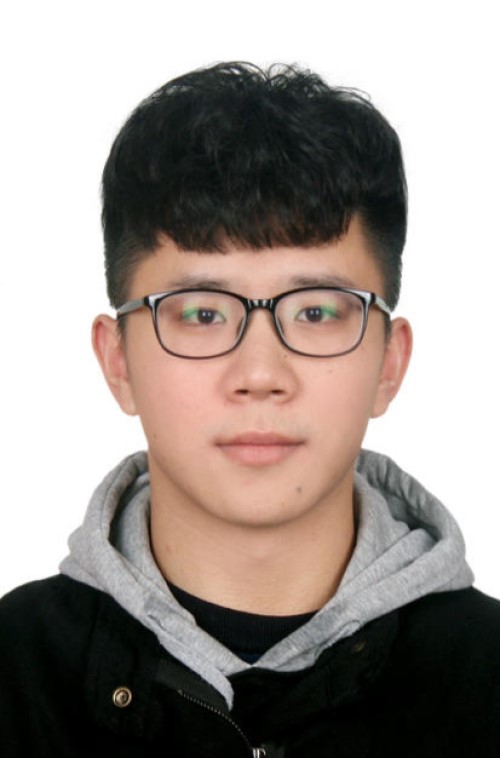}}]
{Zhengxiong Jia} received his B.S. degree in Computer Science from Renmin University of China, Beijing, China, in 2018. He is currently a graduate student at School of Information, Renmin University of China, pursuing his master degree on image captioning.
\end{IEEEbiography}

\begin{IEEEbiography}[{\includegraphics[width=1in,height=1.25in,clip,keepaspectratio]{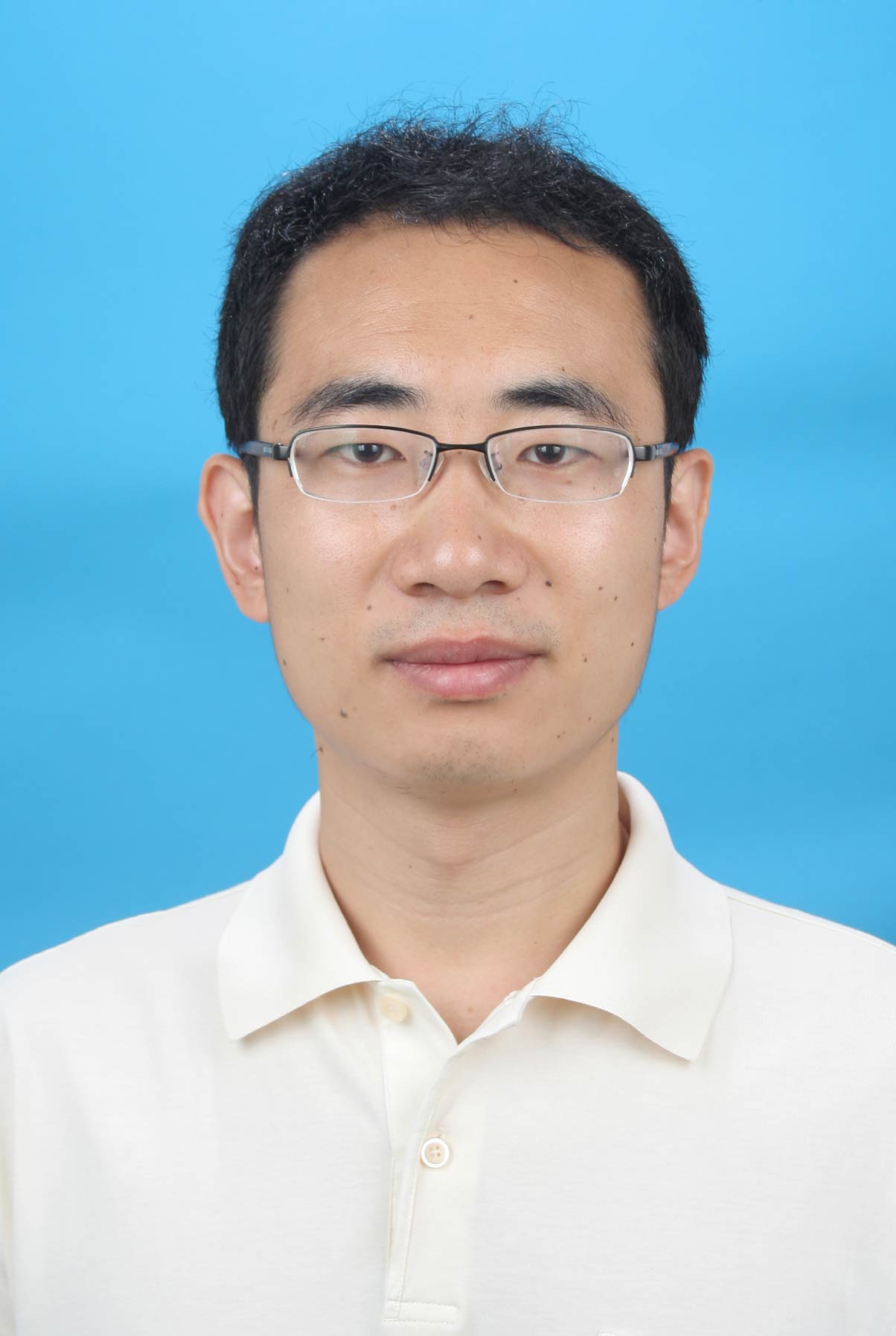}}]
{Gang Yang} received his Ph.D. degree in Innovative Life Science from University of Toyama, Toyama, Japan in 2009. He is currently an Associate Professor at School of Information, Renmin University of China, Beijing, China. His current research interests include computational intelligence, multimedia computing and machine learning.
\end{IEEEbiography}

\begin{IEEEbiography}[{\includegraphics[width=1in,height=1.25in,clip,keepaspectratio]{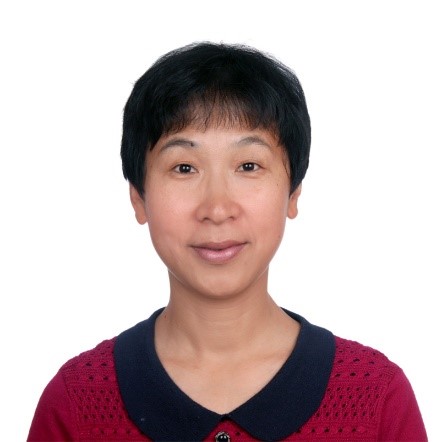}}]
{Jieping Xu} received her Ph.D. degree in Computer Science from the Institute of Acoustics, Chinese Academy of Sciences in 1999. She is currently an Associate Professor at School of Information, Renmin University of China, Beijing, China. Her research interests include multimedia computing and retrieval.
\end{IEEEbiography}

\vfill


\clearpage\end{CJK*}
\end{document}